\documentclass[10pt,twocolumn,letterpaper]{article}

\usepackage{iccv}
\usepackage{times}
\usepackage{epsfig}
\usepackage{graphicx}
\usepackage{amsmath}
\usepackage{amssymb}
\usepackage{booktabs}
\usepackage{multirow}
\usepackage{bbm}
\usepackage{verbatim}
\usepackage{soul}
\usepackage{cancel}
\usepackage{array}
\usepackage{makecell}
\usepackage{subfig}
\usepackage[usenames, dvipsnames]{color}
\usepackage[accsupp]{axessibility}  % Improves PDF readability for those with disabilities.

\newcommand\numberthis{\addtocounter{equation}{1}\tag{\theequation}}

\newcommand{\1}[0]{\mathbbm{1}}
\newcommand{\JXDE}[0]{\textit{JADE} / \textit{JFDE}}
\newcommand{\JADE}[0]{\textit{JADE}}
\newcommand{\JFDE}[0]{\textit{JFDE}}
\newcommand{\XDE}[0]{\textit{ADE} / \textit{FDE}}
\newcommand{\ADE}[0]{\textit{ADE}}
\newcommand{\FDE}[0]{\textit{FDE}}

\usepackage[pagebackref=true,breaklinks=true,letterpaper=true,colorlinks,bookmarks=false]{hyperref}

\newcommand{\deva}[1]{{\leavevmode\color{Red}[Deva: #1]}}
\newcommand{\erica}[1]{{\leavevmode\color{blue}}}

\definecolor{purple}{RGB}{125, 0, 125}

\iccvfinalcopy % *** Uncomment this line for the final submission

\def\iccvPaperID{1202} % *** Enter the ICCV Paper ID here

\ificcvfinal\fi

\begin{document}

%%%%%%%%% TITLE
\title{Joint Metrics Matter: A Better Standard for Trajectory Forecasting }

\author{Erica Weng\thanks{Correspondence to: \tt\small eweng[at]cmu.edu}
\;\;\;\;\;\;\; Hana Hoshino \;\;\;\;\;\;\; Deva Ramanan \;\;\;\;\;\;\; Kris Kitani\\\\
Carnegie Mellon University
}

\maketitle
\ificcvfinal\thispagestyle{empty}\fi

%%%%%%%%% ABSTRACT
\begin{abstract}
Multi-modal trajectory forecasting methods commonly evaluate using single-agent metrics (marginal metrics),
such as minimum Average Displacement Error (ADE) and Final Displacement Error (FDE), which fail to capture joint performance of multiple interacting agents. Only focusing on marginal metrics can lead to unnatural predictions, such as colliding trajectories or diverging trajectories for people who are clearly walking together as a group. Consequently, methods optimized for marginal metrics lead to overly-optimistic estimations of performance, which is detrimental to progress in trajectory forecasting research.
In response to the limitations of marginal metrics, we present the first comprehensive evaluation of state-of-the-art (SOTA) trajectory forecasting methods with respect to multi-agent metrics (joint metrics): JADE, JFDE, and collision rate. We demonstrate the importance of joint metrics as opposed to marginal metrics with quantitative evidence and qualitative examples drawn from the ETH / UCY and Stanford Drone datasets. 
We introduce a new loss function incorporating joint metrics that, when applied to a SOTA trajectory forecasting method, achieves a 7\% improvement in JADE / JFDE on the ETH / UCY datasets with respect to the previous SOTA. Our results also indicate that optimizing for joint metrics naturally leads to an improvement in interaction modeling, as evidenced by a 16\% decrease in mean collision rate on the ETH / UCY datasets with respect to the previous SOTA. Code is available at \texttt{\hyperlink{https://github.com/ericaweng/joint-metrics-matter}{github.com/ericaweng/joint-metrics-matter}}.
\end{abstract}
\vspace{-4mm}

%%%%%%%%% BODY TEXT
% \vspace{-2mm}
\section{Introduction}
% \vspace{-2mm}

\begin{figure}[t]
\begin{center}
  \includegraphics[width=.5\textwidth]{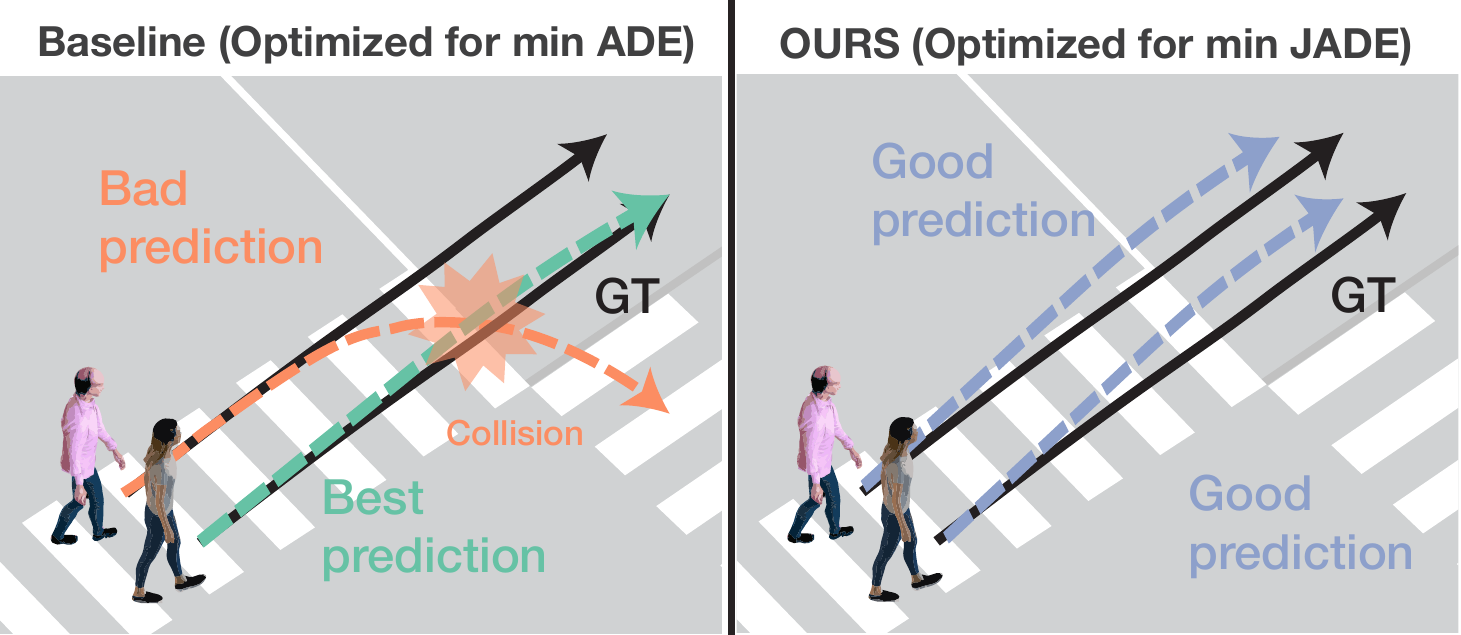}
\end{center}
\caption{Multi-agent trajectory forecasting methods are optimized for single-agent metrics like \ADE{} (left panel). As a result, within a single joint future, the method may predict very good trajectories for some agents (\eg the green agent), but very bad predictions for others (\eg the orange agent). Our Joint AgentFormer, optimized for \textit{JADE} (right panel), encourages the predictions of \textit{all agents} within a joint future to be close to the ground-truth.}
\label{fig:teaser}
\end{figure}

% ==================== Begin Introduction ====================

Because the future is inherently multi-modal, the trajectory forecasting community has adopted the use of \textit{top-K} metrics, average displacement error (\textit{ADE}) and final displacement error (\textit{FDE}), to evaluate and optimize methods.
Under this evaluation, stochastic trajectory forecasting models generate a set of $K$ different possible futures for each target sequence of a certain time length. Each possible future represents a set of \textit{x}-\textit{y} position predictions for all agents in the sequence; we call each set of predictions a ``\textit{sample}."
% By ``predicted future," we mean the predicted future timesteps for a single agent; by ``predicted joint future," we mean the predicted future timesteps for all agents in the sequence.
Min \XDE{} is then calculated by finding, for each agent, the lowest-error prediction across the $K$ samples when compared to the ground-truth. 
% From this perspective, top-$K$ is an optimistic metric since it gives the ``the benefit of the doubt" when selecting the best of $K$ predictions.
Unfortunately, \textit{ADE} / \textit{FDE} do not explicitly account for \textit{interacting} futures. The future is not only multi-modal but also interacting; the future trajectories of each agent depend on one another.
For example, a car merging into a busy road may either choose to push its way in, or wait passively for an opening.
If the car merges, other vehicles on the road will slow down to let it in.
If the car waits, others will maintain their speed.
A scenario where the car merges but the vehicles on the road maintain their speed and crash into the car would not be a likely future.

Most trajectory forecasting methods produce predictions for all agents in the sequence \textit{jointly}, in order to better model these interactive scenarios. However, under standard top-$K$ evaluation, the predictions are considered for each agent \textit{marginally}, and thus the predictions used in evaluation need not come from the same sample.
% This makes top-$K$ evaluation \textit{even more} optimistic.
A method may achieve good \XDE{} simply by producing an accurate prediction for each agent in any one of its $K$ samples, and then mixing-and-matching the best predictions for each agent;
it need not accurately model the joint future within a single sample.

Only considering marginal metrics during evaluation can result in an overly optimistic estimation of performance. A method may achieve excellent \XDE{} yet still fail at joint interaction modeling, such as predicting collision-free trajectories,
as seen in Figure \ref{fig:teaser}. 
After crossing the crosswalk, it is reasonable that a pedestrian either continues straight down the sidewalk, turns left onto the sidewalk, 
or turns right to cross the other crosswalk. 
Supposing all three possibilities are equally likely and well-represented by the ground truth data,
a model that has been trained relying on standard min \XDE{} metrics
(\textit{e.g.} that shown in left panel of figure) may predict any one of these three futures. %the orange trajectory turning right to cross the other crosswalk for the pink pedestrian,
%and the green trajectory going straight for the gray pedestrian,
%as turning right and going straight are both equally likely futures. 
However, since the method is optimized for each agent individually, 
it may not enforce the correct {\em group} behavior;
%consider that if both pedestrians are walking grouped, 
two pedestrians walking side-by-side are likely to choose the same mode.

A work from the vehicle forecasting community \cite{sun_scalability_2020} introduced \textit{joint metrics}, which they called ``scene-level" \ADE{} and \textit{FDE}. These metrics, rather than mix-and-match the minimum \XDE{} prediction per-agent, instead average over all agents within a sample before taking the minimum-error sample as the one that ``counts." Performing well with respect to joint metrics necessitates not only predicting accurate marginal futures, but also accurate joint futures. Joint metrics are also reported in the Waymo Motion Prediction challenge and several other vehicle forecasting works; however, they have failed to gain traction in pedestrian forecasting evaluation, and few if any methods that evaluate on pedestrian datasets report them.
% joint ``scene-level" metrics
%\erica{should i introduce JADE / JFDE here?}
% (\textit{JADE} and \textit{JFDE}, or \JXDE{} for short) 

Diverging from the names used by \cite{sun_scalability_2020}, we shall instead refer to ``scene-level" \ADE{} / \FDE{} as ``joint" \XDE{} (\textit{JADE} / \textit{JFDE}) in order to avoid confusion with the overloaded term ``scene," which is also used in trajectory forecasting literature to refer to environment information such as semantic maps or the location of a dataset. We define these metrics and compare the differences with their marginal counterpart \XDE{} in Section \ref{sec:metrics}. 

In theory, current forecasting architectures may already be capable of modeling joint futures and multi-agent interactions through the use of graph and attention-based architectures such as GNNs and Transformers.
These architectures are designed to explicitly model social interactions between agents so as to predict realistic joint futures,
and have led to great improvements in \ADE{} / \textit{FDE}. 
% This is in spite of increasing studies showing that adversarial attacks can exploit the fact that current forecasting methods are optimized for marginal metrics \cite{saadatnejad_are_2021, Cao2022-hq, Zhang2022-dp}. 
However, increasing studies show that adversarial attacks can cause methods to produce unrealistic joint futures and poor results~\cite{saadatnejad_are_2021, Cao2022-hq, Zhang2022-dp}. For example, \cite{saadatnejad_are_2021} showed that ``socially-aware" models may not be as socially-aware as they claim to be, because well-placed attacks can cause predictions with colliding trajectories.
To more realistically assess the performance of multi-agent forecasting, we advocate for the use of joint metrics over marginal metrics in method \textit{evaluation}.
% and even trajectories with poor ADE / FDE.
Furthermore, we hypothesize that multi-agent architectures fall short in modeling realistic agent interactions because they are optimized with respect to only marginal losses (that are driven by the field's focus on marginal metrics). To test this hypothesis, we modify the loss function on a popular SOTA method to include a joint loss term.
We show with this simple and straightforward modification, SOTA methods become far more accurate at modeling multi-agent interactions. %adding joint losses allows methods to forecast more realistic joint futures.

Our contribution is thus two-fold: 
\begin{itemize}
    % \vspace{-2mm}
    \item We present the first comprehensive evaluation of SOTA trajectory forecasting methods with respect to \textit{joint metrics}: \textit{JADE}, \textit{JFDE}, and collision rate. We show that \textit{ADE} / \textit{FDE} are an overestimation of performance on joint agent behavior; \JXDE{} is often 2x worse than the reported \XDE{} across methods for $K=20$ samples. We illustrate the importance of joint metrics over marginal metrics with quantitative evidence as well as qualitative examples drawn from SOTA methods on the ETH~\cite{pellegrini_youll_2009} / UCY~\cite{lerner2007crowds} and SDD~\cite{robicquet_learning_2016} datasets.
    % \vspace{-2mm}
    \item We introduce a simple multi-agent {\em joint loss} function that, when used to optimize a SOTA trajectory forecasting method, achieves a improvement of 7\% over the previous SOTA with respect to the joint metrics \JXDE{} on the ETH / UCY datasets. We show that this lower \JXDE{} is also linked with a 16\% reduction in collision rate as compared with the previous SOTA, substantiating the hypothesis that optimizing with respect to joint metrics also improves collision-avoidance modeling.
\end{itemize}

% \vspace{-2mm}
\section{Related Work}
% \vspace{-2mm}
\noindent\textbf{Joint Evaluation Metrics.}
\JXDE{} was introduced in \cite{sun_scalability_2020}, 
but few papers in either the vehicle or pedestrian forecasting have evaluated and 
reported performance on it. As far as we know, the only papers that report on it are \cite{sun_scalability_2020,Ettinger2021-yl,Sun2022-mj}, all from the vehicle forecasting community. 
In pedestrian forecasting, \JXDE{} is overlooked in favor of \XDE{}.

Collision Rate (\textit{CR}) is a joint evaluation metric that has seen greater attention in more recent works as members of the trajectory 
forecasting community begin to pay more attention to social compliance and effective joint modeling~\cite{Kothari2022-hl, sohn_a2x_2021,Kothari2022-cy, kothari_interpretable_2021, liu_social_2021, saadatnejad_are_2021}.

In the autonomous vehicle setup, other joint evaluation metrics have been proposed, 
including Driveable Area Compliance~\cite{chang_argoverse_2019}, Miss Rate~\cite{chang_argoverse_2019, sun_scalability_2020, Ettinger2021-yl}, 
and mean Average Precision~\cite{sun_scalability_2020, Ettinger2021-yl},
the latter two taking inspiration from the object detection evaluation setup.
These metrics all measure ``realism" aspects of predicted trajectories, but have still not become widespread in the pedestrian trajectory forecasting space.

The NLL metric captures joint-agent performance because it averages over all pedestrians $N$, but it is insufficient for capturing best-of-$K$ multi-modality because it averages over samples instead of taking the minimum:
% \vspace{-3mm}
$$NLL = \frac{1}{KNT} \sum_{k,t,i} -\log KDE\_PDF(p_{t,n}^{(i)}, p_{t,n}^{*(i)})$$ 
% \vspace{-5mm}

\noindent Thus, when used to report performance on real datasets with only one ground truth future mode, NLL does not reward plausible but unrepresented modes.

Precision / Recall can only be calculated on datasets that report multiple ground truth modes, such as the Forking Paths Dataset~\cite{liang_garden_2020}, so that is a limitation.
% Marginal Multi-modal Evaluation Metrics include NLL, AMD

% Perhaps this is in part because these metrics are not as relevant to humans as they are to vehicles. For example, 
% DAC does not apply as well to pedestrians because humans are not as limited as vehicles in the spaces they can feasibly occupy in outdoor environments.
% Miss rate and mAP for humans
% \vspace{2mm}

\noindent\textbf{Multi-modal Joint Trajectory Forecasting.}
\label{sec:multimodal_tf}
Modern trajectory forecasting methods typically learn to predict trajectory distributions using deep models trained to imitate real datasets. In order to capture multiple possible futures, many methods are based on deep generative architectures~\cite{kingma2013auto, goodfellow2014generative, rezende2015variational}, such as conditional variational autoencoders (CVAEs)~\cite{lee2017desire,yuan2019diverse,ivanovic2019trajectron,tang2019multiple,weng2020joint}, generative adversarial networks (GANs)~\cite{gupta2018social,sadeghian2019sophie,kosaraju2019social,zhao2019multi,Kothari2022-cy}, and normalizing flows (NFs)~\cite{rhinehart2018r2p2,rhinehart2019precog,guan2020generative}.
%multi-modal futures.
To model the joint behavior of multiple socially interacting agents, 
many methods have made use of graph-based computation structures with attention and pooling 
layers such as GNNs and transformers~\cite{ivanovic2019trajectron, mangalam_it_2020, salzmann_trajectron_2021, kipf2018neural,
alet_neural_2019,Liang2020-of, chen_relational_2020,su_trajectory_2022, mohamed_social-stgcnn_2020,
Yu2020-df,Yuan2020-vw,Yuan2021-tp, Amirloo2022-se, Cao2022-uv, bahari_svg-net_2021,Huang2021-sv}. SOTA methods have employed a variety of techniques for better modeling, such as using waypoints and goals to break the forecasting problem into a series of hierarchical steps~\cite{Chiara2022-xa, mangalam_goals_2020, zhao_tnt_2020,chai_multipath_2019, chai2020multipath, Yue2022-os, Zhao2021-ca, Yao2021-xn, Wong2022-kx}, mathematical and physical modeling techniques such as social forces~\cite{helbing_social_1995} alongside deep architectures~\cite{Yue2022-os, Wong2022-kx, liu_energy-based_2021}, and other paradigms and techniques such as memory retrieval~\cite{Xu2022-ki}, contrastive learning~\cite{liu_social_2021}, and causal disentanglement~\cite{liu_towards_2021}. Despite the large improvements made with respect to \XDE{}{}, progress with respect to \JXDE{} is still unknown, as none of the aforementioned methods evaluate with respect to \JXDE{}.

\erica{this section can probably be expanded a bit more. I'm not sure if I should add a discussion of multi-modal metrics (NLL, AMD, etc.) since they aren't particularly relevant to our current paper. Perhaps I can talk a bit but not too much about how adversarial explorations have revealed that socially-aware models are not really socially-aware, as I mentioned in the Introduction.}

% \vspace{-4mm}
\section{Metrics Definitions}
\label{sec:metrics}
% \vspace{-2mm}
We formulate the multi-agent trajectory forecasting problem as predicting the future trajectories of $N$ agents conditioned on their past trajectories. For observed history timesteps $t \le 0$, we represent the state for agent $n$ at timestep $t$ as $x_{t,n} \in \mathcal{\mathbb{R}}_d$, which includes the position, velocity, and (in some methods) the heading angle of the agent. We denote the joint observation history for all $N$ agents over all $T$ timesteps as $\textbf{x} = (x_{1,1},\dots, x_{T,N})$.
% We denote the history of all agents as X = (
% X − H , X − H +1,...,X 0) which  includes  the  joint agent state at all H + 1 observed timesteps. 
For future timesteps $t>0$, we represent the ground-truth positions of agent $n$ at timestep $t$ as $y^*_{t, n} \in \mathcal{\mathbb{R}}_2$, which includes a 2D \textit{x}-\textit{y} position. 
We denote the ground-truth trajectories of over all agent-timesteps as $\textbf{y}^* = (y^*_{1,1}, \dots, y^*_{T,N})$. 
Similarly, we represent the joint position predictions for agent $n$ at timestep $t$ in prediction sample $k$ as $y_{t, n}^{(k)} \in \mathcal{\mathbb{R}}_2$, and the position predictions over all agent-timestep-samples as $\textbf{y}$.

% \subsection{Metrics}
We define the marginal and joint metrics used in evaluation.
The standard used in evaluations is $8$ history or observation timesteps and $T=12$ future timesteps for a total of $20$ frames per sequence sampled at $2.5$ fps; thus $3.2s$ of history observation and $T=4.8 s$ of future. $K$ is the number of samples, or possible futures, produced by the model for a single $20$-frame sequence; the standard used in evaluations is $K=20$. $N$ is the number of agents, which varies by sequence.

% \vspace{2mm}
% \noindent\textbf{Top-$K$ Minimum Displacement Error (XDE).}
\noindent\textbf{Marginal Metrics (\XDE{}).}
Throughout the paper we use \XDE{} to refer to \textit{top-$K$ minimum error} rather than average error, as this is the standard notation used in multi-modal human trajectory forecasting evaluation.
\begin{equation}
 ADE(\textbf{y}, \textbf{y}^*) = \frac{1}{TN}\sum_{n=1}^N \min_{k=1}^K \sum_{t=1}^T \Big\| y_{t,n}^{(k)}-y_{t,n}^*\Big\|_2^2
\end{equation}
\begin{equation}
FDE(\textbf{y}, \textbf{y}^*) = \frac{1}{N}\sum_{n=1}^N \min_{k=1}^K \Big\| y_{T,n}^{(k)}-y_{T,n}^*\Big\|_2^2
\end{equation}
% \vspace{2mm}\noindent\textbf{Top-$K$ Minimum Scene-level 
% \vspace{-2mm}

\noindent\textbf{Joint Metrics (\JXDE{}).} The difference between the calculation of \XDE{} and \JXDE{} is small but significant: swapping the order of taking the minimum over samples $k$ and taking the average over agents $n$. This difference means we cannot mix-and-match agents between different samples; rather we must take the average error over all agents within a sample before we select the best one to use in performance evaluation.
% \vspace{-2mm}
\begin{equation}
 JADE(\textbf{y}, \textbf{y}^*) = \frac{1}{TN} \min_{k=1}^K \sum_{n=1}^N\sum_{t=1}^T \Big\| y_{t,n}^{(k)}-y_{t,n}^*\Big\|_2^2
\end{equation}
\begin{equation}
JFDE(\textbf{y}, \textbf{y}^*) = \frac{1}{N}\min_{k=1}^K \sum_{n=1}^N \Big\| y_{T,n}^{(k)}-y_{T,n}^*\Big\|_2^2
\end{equation}
\noindent\textbf{Collision Rate (\textit{CR}).}
Collision Rate is the proportion of agents whose path intersects in time and space (by a certain threshold) with at least one other predicted agent future in a $T=12$-frame prediction sample.

If we adopt top-$K$ evaluation and use the minimum-\JADE{} sample as an optimistic measure of the method's prediction ability, then the collision rate of that sample provides an optimistic estimate of the model's collision-avoidance ability. Thus, we define $CR_{JADE}$, the collision rate of the minimum-\JADE{} sample:
% \vspace{-2mm}
\begin{align*}
    CR_{JADE}(\textbf{y}) &= \frac{1}{N} \sum_{n=1}^N \1\Big[collision\big(\textbf{y}_{n}^{(k)}, \textbf{y}_{m\ne n}^{(k)}\big)\Big]; \numberthis \label{eq:CR_JADE} \\
    \;\;\;&k = {\arg\min}_{k'} \sum_{n=1}^N\sum_{t=1}^T \Big\| y_{t,n}^{(k')}-y_{t,n}^*\Big\|_2^2
\end{align*}
\noindent where $\textbf{y}_{n}^{(k)}$ represents all predictions $y_{t,n}^{(k)}$ for $t=[1,...,T]$, and $collision$ is a function that returns True if any two line segments formed by $\big(y_{t,n}^{(k)}, y_{t+1, n}^{(k)}\big) \in \textbf{y}_{n}^{(k)}$ and $\big(y_{t,m}^{(k)}, y_{t+1, m}^{(k)}\big) \in \textbf{y}_{m}^{(k)}\forall\; m \ne n$  come within $2b$ of each other, and False otherwise. For our evaluations, we use an agent radius of $b=0.1$ meters, as used in \cite{kothari_human_2021}.
Adversarial examples discounted, $CR_{JADE}$ should be lower than $CR_{mean}$, the mean collision rate across all samples. This is because the sample with the best \JADE{} is that which is closest to the ground-truth, which has few collisions, as seen in the last row of Table \ref{table:all_CR_of_best-JADE_sample_mean_collision_rate}.

Because proper learning of social interactions should result in low collision rate, we would ideally like all samples produced by a model to avoid collisions. Thus, we also define $CR_{mean}$ as defined in past work \cite{kothari_human_2021, sohn_a2x_2021}:
% \deva{Does anyone else report mean CR? If so, we should cite and could use their justification - e.g., "as in past work..." Since you also report $CR_{SDE}$, I would start by defining that, which is more natural given the flow. You can state "Because we would like all joint predictions from a model to avoid collisions, one can also compute $CR_{mean}$."} 
% \vspace{-3mm}
\begin{align}
    CR_{mean}(\textbf{y}) &= \frac{1}{NK} \sum_{k=1}^K \sum_{n=1}^N \1\Big[collision\big(\textbf{y}_{n}^{(k)}, \textbf{y}_{m\ne n}^{(k)}\big)\Big]
\end{align}
As it considers the mean over samples, rather than the min as in top-$K$ evaluation, $CR_{mean}$ provides a holistic rather than optimistic evaluation of a model's collision-avoidance ability.

% \vspace{-2mm}
\section{Joint Optimization of SOTA Methods}
% \vspace{-2mm}
\noindent We introduce an improvement upon AgentFormer \cite{yuan_agentformer_2021} and View Vertically \cite{Wong2022-kx} that improves performance with respect to \JXDE{} and, in the case of AgentFormer, Collision Rate. 
We selected two trajectory prediction models to show our methods' broad applicability.

\subsection{AgentFormer}
The AgentFormer architecture is a CVAE structure with transformer layers that factorizes the trajectory forecasting problem into a prediction over a latent space of possible futures $\textbf{z}$ conditional on the trajectory histories $\textbf{x}$, and a prediction over decoded possible futures $\textbf{y}$ conditional on the latent $\textbf{z}$. We make no modifications to the AgentFormer architecture, and thus refer readers to the original paper \cite{yuan_agentformer_2021} for more detail.

% minimizing the ELBO loss.
% given the latent while minimizing the KL Divergence is Gaussian
% which assuming the underlying distribution over latent futures to be Gaussian, breaks down into an L2 displacement error term. 
% given the trajectory history and the latent
% This multiple choice loss \cite{} over predicted futures is equivalent to an ADE loss.
% , giving additional weight to the best predicted sample in the training signal.
% AgentFormer utilizes 
AgentFormer training involves a two-step procedure: a first step to learn accurate trajectory decoding, and a second tuning step to learn to produce diverse prediction samples. 
During the first step, AgentFormer makes use of the negative evidence lower bound (ELBO) loss function to encourage the CVAE model's predictions to match the ground-truth positions while maintaining that the latent space of possible futures adhere to the Gaussian distribution.
\begin{equation}
\label{eq:elbo}
    \mathcal{L}_{elbo} = - \mathop{\mathbb{E}_{q_\phi}}[\log p_{\theta} (\textbf{y}|\textbf{z}, \textbf{x}) ]
 + \text{KL} (q_{\phi}(\textbf{z}|\textbf{y}, \textbf{x}) || p_{\theta}(\textbf{z}|\textbf{x})]
\end{equation}
This loss can be rewritten as the three terms in black: % in equation \ref{eq:elbo_pre}. 
\begin{align*}
    \mathcal{L}_{elbo} &= \sum_n ||\textbf{y}_n^{(0)}-\textbf{y}^*||^2 %\\
    + \sum_n \min_{k=1}^K ||\textbf{y}_n^{(k)}-\textbf{y}_n^*||^2   \numberthis \label{eq:elbo_pre}\\
    & \textcolor{blue}{+  \min_{k=1}^K \sum_n 
    ||\textbf{y}_n^{(k)}-\textbf{y}_n^*||^2}
    % + \text{KL}\big(q_{\phi}(\textbf{z} |\textbf{y}, \textbf{x})\;
    % \big|\big|\; 
    % p_\theta ({\textbf{z} | \textbf{x}})\big)
    + \text{KL}(q_{\phi}(\textbf{z}|\textbf{y},\textbf{x})\;||\; p_\theta (\textbf{z}|\textbf{x}) )
\end{align*}
The first and second terms are reconstruction terms resulting from the first term of equation \ref{eq:elbo} (the ELBO likelihood term).
% of $\mathcal{L}_{elbo}$). 
The first term is a general reconstruction loss that encourages a single predicted future $\textbf{y}^{(0)}$ to be close to the ground-truth $\textbf{y}^*$. The second term is a marginal sample reconstruction loss that encourages the min \textit{ADE} prediction for agent $n$, 
$\min_k ||\textbf{y}_n^{(k)} - \textbf{y}_n^* ||$, 
to be close to the ground-truth $\textbf{y}_n^*$. 
\erica{Maybe I should ignore the first term (reconstruction) since it might be confusing how it differs from the second term (sample) term?}
The fourth term is equivalent to the second term of equation \ref{eq:elbo} (the KL divergence term), which encourages the \textit{CVAE Prior} network, $p_\theta$, to learn the latent distribution encoded by the posterior network $q_\phi$.

% As the model produces multiple futures to maximize the probability that the best of $K$ predicted futures per agent matches the ground-truth future. There is also a reconstruction objective, which is to maximize the probability that the highest-likelihood future predicted by the model matches the ground-truth future.
In our joint optimization, we modify the objective function by adding the third term in blue, a joint sample reconstruction loss that encourages the min \textit{JADE} prediction
$\min_k \sum_n ||\textbf{y}_n^{(k)} - \textbf{y}_n^* ||$ 
to be close to the ground-truth $\textbf{y}^*$. 
% This is equivalent to adding a to the likelihood term of $\mathcal{L}_{elbo}$.

While the first training step learns a latent space that maximizes the probability that decoded predictions match the ground-truth, AgentFormer makes use of a second training step that is designed to encourage the decoded predictions to be diverse. Here, the CVAE weights are fixed, and the \textit{CVAE Prior} network's latent sampler is swapped out for a \textit{DLow Trajectory Sampler}. This new sampler module learns a fixed set of $K$ linear transformations of the latent $\textbf{z}$ that are trained to be different from one another via a DLow diversity loss $\mathcal{L}_{samp}$ \cite{yuan2020dlow}.
We signify the new prior network with the \textit{DLow Trajectory Sampler} as $r_\theta$. 
In our joint optimization, we modify AgentFormer's objective function by adding a joint sample reconstruction term, just as we did in the first training step. The final objective function for the second training step is therefore:
% diversity-promoting loss proposed in \cite{yuan2020dlow}, which has the second and fourth terms of equation \ref{eq:elbo_obj}, 
\begin{align*}
    % \mathcal{L}_{samp} &= \textcolor{red}{\cancel{ \sum_n  \min_{k=1}^K  ||\textbf{y}_n^{(k)}-\textbf{y}_n^*||^2} } \textbf{}
    \mathcal{L}_{samp} &= \sum_n 
    \min_{k=1}^K  ||\textbf{y}_n^{(k)}-\textbf{y}_n^*||^2
    + \textcolor{blue}{\min_{k=1}^K \sum_n 
    ||\textbf{y}_n^{(k)}-\textbf{y}_n^*||^2} \\
    & + \text{KL}\big(r_{\theta}(\textbf{z}|\textbf{x})\;||\; p_\theta(\textbf{z}|\textbf{x})\big) \numberthis \label{eq:elbo_dlow} \\
    & + \frac{1}{K(K-1)}\sum_{k_1=1}^K\sum_{k_1\ne k_2}^K \exp\Bigg( - \frac{||\textbf{y}^{(k_1)} - \textbf{y}^{(k_2)} ||}{\sigma_d} \Bigg) 
\end{align*}

\noindent The first term is the marginal sample reconstruction loss, analogous to the second term of Equation \ref{eq:elbo_pre}. The blue second term, our addition, is equivalent to the joint sample reconstruction loss we added to the first training step. The third term is a KL term which encourages the new prior network $r_\theta$ (with the new \textit{Trajectory Sampler} module), to be near the distribution of the original prior network $p_\theta$ (with the old \textit{CVAE Prior} sampler, which was learned in the first training step). The fourth term is the diversity loss, which encourages the fixed set of $K$ futures to be diverse from one another.
\subsection{View Vertically}
View Vertically is a simple hierarchical method with two modules: a coarse-level waypoint prediction module, and a fine-level trajectory prediction module. The coarse-level module forecasts the future waypoints in the spectrum domain, and the fine-level module interpolates waypoints in the spectrum domain, and then decodes full trajectories in coordinate space. We make no modifications to the architecture, and thus refer readers to the original paper \cite{Wong2022-kx} for more detail.

Each module is trained independently. The coarse-level module is optimized according to the loss written in black:

\begin{align*}
    \mathcal{L}_{coarse} &= \frac{1}{N_{way}N_{agents}} %\Bigg( 
    \sum_{n} \min_{k=1}^K \sum_{m=1}^{N_{way}}   ||\textbf{y}_{m,n}^{(k)}-\textbf{y}_{m,n}^*||^2  %\\
     \numberthis \label{eq:vv}\\
    & \textcolor{blue}{+\; \omega \cdot \frac{1}{N_{way}N_{agents}} 
    \min_{k=1}^K \sum_{n} \sum_{m=1}^{N_{way}} 
    ||\textbf{y}_{m,n}^{(k)}-\textbf{y}_{m,n}^*||^2 }
    % \Bigg)
\end{align*}

Here, $N_{way}$ is the set of waypoint timesteps used for coarse prediction, optimized as a hyperparameter. Similar to how we do with AgentFormer, to optimized View Vertically with joint metrics, we add the blue second term, the joint optimization term. Here, $\omega$ is a weighting hyperparameter that balances how much the to consider the marginal term vs. the joint term.

The fine-level module is optimized according to the loss written in black:

\begin{align*}
    \mathcal{L}_{fine} &= \frac{1}{TN} %\Bigg( 
    \sum_{n} \min_{k=1}^K \sum_{t}   ||\textbf{y}_{t,n}^{(k)}-\textbf{y}_{t,n}^*||^2  %\\
     \numberthis \label{eq:vv}\\
    & \textcolor{blue}{+\; \omega \cdot \frac{1}{TN} 
    \min_{k=1}^K \sum_{n} \sum_{t} 
    ||\textbf{y}_{t,n}^{(k)}-\textbf{y}_{t,n}^*||^2 }
    % \Bigg)
\end{align*}

We add in the term in blue, analogous to the term we add for the coarse-level module.

% \vspace{-2mm} 
\section{Evaluation Setup}
% \vspace{-2mm}

\noindent \textbf{Datasets.} We evaluate on the commonly-used pedestrian trajectory datasets ETH~\cite{pellegrini_youll_2009} / UCY~\cite{lerner_crowds_2007}, used across the field in nearly all pedestrian trajectory forecasting works. ETH / UCY features bird's eye views trajectories of 
pedestrians in 5 different environment scenes given in real-world coordinates (meters). Following 
\cite{alahi_social_2016,gupta_social_2018,yuan_agentformer_2021,mangalam_goals_2020,mangalam_it_2020,salzmann_trajectron_2021}, we use the leave-one-out training and evaluation setup, where we train on all but one of the 5 environment scenes, and evaluate on the left-out scene. 

We also evaluate on the Stanford Drone 
Dataset (SDD)~\cite{robicquet_learning_2016}, which features trajectories captured via drone from a bird’s eye viewpoint across 20 different scenes on Stanford University's campus given in image pixel coordinates. Instead of using the raw data, we chose to use the TrajNet split~\cite{Kothari2022-hl} of SDD, a downsampled and smaller subset of the original data considering only pedestrian trajectories, following the training and evaluation setup of
\cite{salzmann_trajectron_2021,sadeghian2019sophie}.
% which is also widely evaluated upon.
% , consisting of trajectories collected across 8 university campus scenes.

\vspace{2mm}
\noindent\textbf{Baselines.}
We compare our approach with current state-of-the-art methods: View Vertically~\cite{Wong2022-kx}, MemoNet~\cite{Xu2022-ki}, Y-Net~\cite{mangalam_goals_2020}, and AgentFormer~\cite{yuan_agentformer_2021}; and other standard baselines from recent years: Trajectron++~\cite{salzmann_trajectron_2021}, PECNet~\cite{mangalam_it_2020}, and S-GAN~\cite{gupta_social_2018}. We re-evaluate using pre-trained models if provided, or retrain and re-evaluate using available source code if not.
% We evaluated SOTA methods from recent years with respect to scene-level metrics JADE / JFDE (table \ref{table:all_min_JADE_min_JFDE}) and collision rate (table \ref{table:all_CR_of_best-JADE_sample_mean_collision_rate}), and include ADE / FDE \ref{table:all_min_ADE_min_FDE} for completeness.
For fair comparison between methods, we provide the model only ground-truth trajectory 
histories, and do not provide scene context information such as images or semantic maps. 
% All methods we evaluate upon either do not accept semantic maps as input, or map function without;  
Y-Net is the only baseline in our evaluation that used semantic map information in the original work~\cite{mangalam_it_2020}; the reason the results we report for Y-Net are not as performant as those reported in their paper is because we retrain without using map information. 
There are a few other reasons for differences between our results and that reported in the original papers:
the numbers originally reported in Trajectron++ were incorrect due to a 
% \hyperlink{https://github.com/StanfordASL/Trajectron-plus-plus/issues/53}{
data snooping bug;
% }; 
thus we re-train and re-evaluate on a corrected version of the code. 
View Vertically used 
% \hyperlink{https://github.com/cocoon2wong/Vertical/issues/1}{
a different sample rate
% } 
on the \texttt{eth} scene and a slightly different split of the \texttt{hotel} scene in the ETH dataset;
thus, we retrain and re-evaluate it on the standard split used by most other methods (\cite{salzmann_trajectron_2021, mangalam_it_2020, mangalam_goals_2020, Xu2022-ki, gupta_social_2018, yuan_agentformer_2021}).
To ensure fair comparison across methods on SDD, we re-train and re-evaluate methods which either used a different split instead of the TrajNet~\cite{kothari_human_2021} split
% PECNet, Y-Net, and \textcolor{red}{MemoNet} originally filtered out non-pedestrians;
(View Vertically used the original raw dataset and downsampled / preprocessed the data themselves) or did not originally train and report results on SDD (AgentFormer and Trajectron++).

% Most current SOTA methods predict the future trajectories of the agent jointly, \ie in the context of its social interactions with other agents

\section{Results and Analyses}
% \vspace{-2mm}

% =============== BEGIN Main Qualitative figure ===================================
\begin{figure*}[t]
\begin{center}
  \includegraphics[width=.95\linewidth]{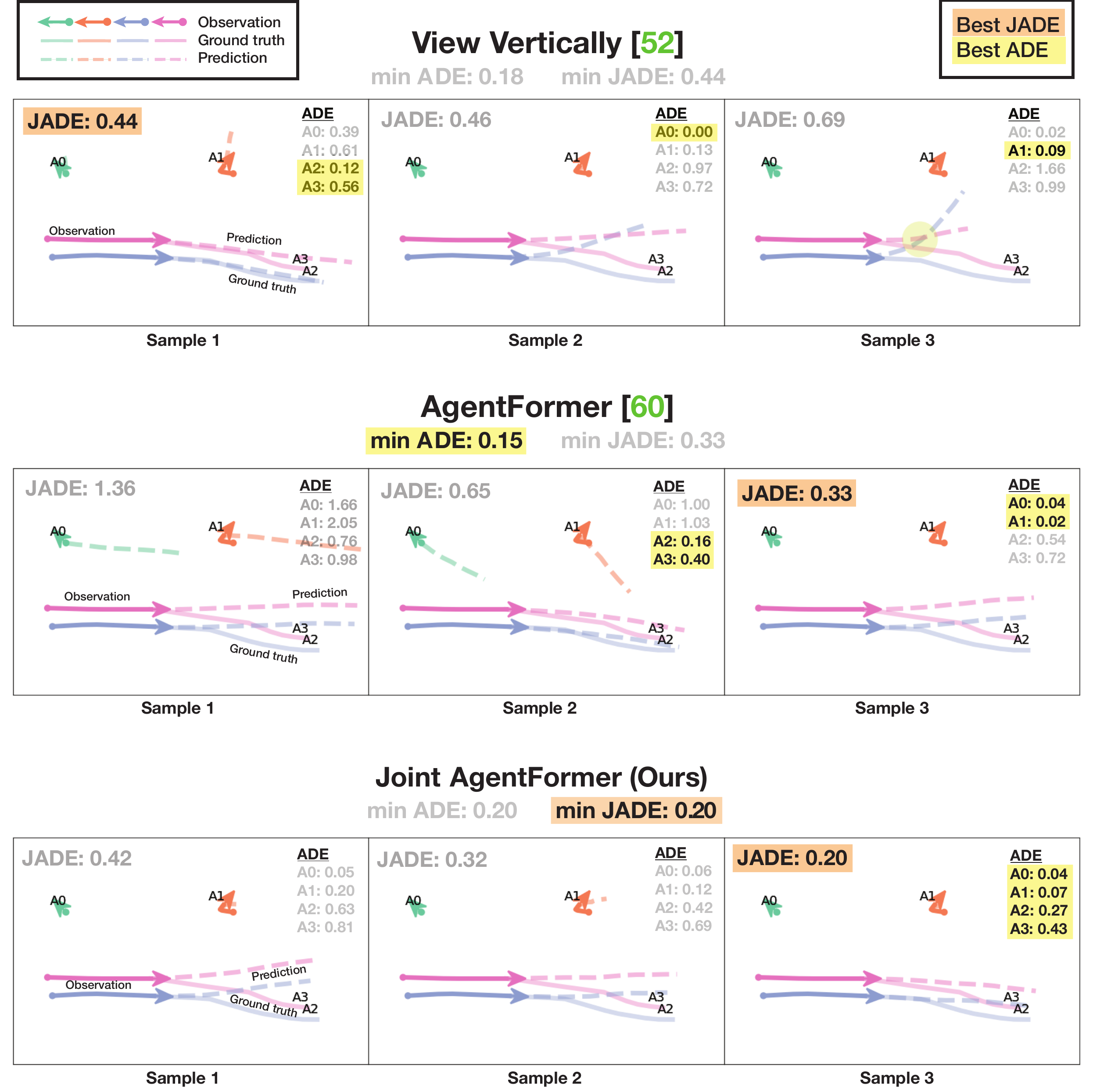}
\end{center}
\caption{
Comparison of Our Joint AgentFormer (last row) with two baselines, View Vertically~\cite{Wong2022-kx} and AgentFormer~\cite{yuan_agentformer_2021}. Legend is in the upper-left corner. A\# stands for Agent \#. Best per-pedestrian \ADE{} values are highlighted yellow; best \JADE{} values are highlighted orange.
View Vertically (1st row) and AgentFormer (2nd row), optimized for \textit{ADE}, achieve a better \ADE{} than Our Joint AgentFormer by mixing and matching pedestrians from different samples. However, they have lower \JADE{} than Our Joint AgentFormer, because no single sample has good \textit{JADE}. On the other hand, our method's best \ADE{} is equal to our best \JADE{} (bottom right), since our method was optimized to encourage all pedestrians within a sample to have low error.
Baseline methods that optimize for \XDE{} rather than \JXDE{} have several other shortcomings. For example, in the top-right panel, View Vertically predicts colliding trajectories (collision denoted by the light yellow circle). In the top-middle panel, it predicts the diverging trajectories for two pedestrians that were clearly walking as a group. In spite of these failures, View Vertically still achieves excellent \textit{ADE}, pointing to a shortcoming in evaluation using only marginal metrics as well optimization using only marginal loss.
}
\label{fig:traj_viz}
\end{figure*}

% =============== BEGIN Main baseline tables ===========================
\setlength{\tabcolsep}{4pt}
\begin{table*}[ht]
\caption{Baseline evaluations across SOTA baselines (first 7 rows) as well as Our Joint AgentFormer method (last row), on the ETH / UCY datasets (first 6 columns) and the Stanford Drone dataset (last column). The metrics being reported (a.) \textit{JADE} / \textit{JFDE}, (b.) $CR_{JADE}$ / $CR_{mean}$, and (c.) \textit{ADE} / \textit{FDE}) are shown in the title of each table. Sequence density, as given by mean number of agents per 20-frame sequence, is shown in parentheses next to each dataset name. Lower values are better; bolded values show best result, underlined values show the second-best. All results are for $K=20$ prediction samples per sequence.}
\label{table:all_min_JADE_min_JFDE_all}
\subfloat[$\min \textit{JADE} / \textit{JFDE}$]{
\label{table:all_min_JADE_min_JFDE}
\resizebox{\textwidth}{!}{
\begin{tabular}{c|ccccc|c||c}
\hline
\noalign{\smallskip}
 & \multicolumn{7}{c}{$\min \text{JADE}_{20} / \text{JFDE}_{20} \downarrow$  (m),  $K=20$ samples} \\
\cline{2-8}
\noalign{\smallskip} 
\;\;\;\;\;\;\;\; \text{Dataset \footnotesize(mean \# peds)} $\rightarrow$ & ETH \footnotesize(1.4) & HOTEL \footnotesize(2.7) & UNIV \footnotesize(25.7) & ZARA1 \footnotesize(3.3) & ZARA2 \footnotesize(5.9) & ETH / UCY Avg. & SDD TrajNet \footnotesize(1.5)\\
\noalign{\smallskip}
\hline
\noalign{\smallskip}
S-GAN~\cite{gupta_social_2018} & 0.919 / 1.742 & 0.480 / 0.950 & 0.744 / 1.573 & 0.438 / 1.001 & 0.362 / 0.794 & 0.589 / 1.212 & 13.76 / 24.84\\
Trajectron++ \cite{salzmann_trajectron_2021} & 0.726 / 1.299 & 0.237 / 0.418 & 0.609 / 1.316 & 0.359 / 0.712 & 0.294 / 0.625 & 0.445 / 0.874 & 11.36 / 18.21\\
PECNet \cite{mangalam2020not} & 0.618 / 1.097 & 0.291 / 0.587 & 0.666 / 1.417 & 0.408 / 0.896 & 0.372 / 0.840 & 0.471 / 0.967 & 10.82 / 19.48\\
Y-Net \cite{mangalam_goals_2020} & 0.495 / \underline{0.781} & 0.205 / 0.386 & 0.695 / 1.559 & 0.487 / 1.045 & 0.492 / 1.101 & 0.475 / 0.974 & 9.67 / \textbf{16.01}\\
MemoNet \cite{Xu2022-ki} & 0.499 / 0.859 & 0.222 / 0.416 & 0.686 / 1.466 & 0.349 / 0.723 & 0.385 / 0.864 & 0.428 / 0.866 & \underline{9.59} / \underline{16.43}\\
View Vertically \cite{Wong2022-kx} & 0.561 / \textbf{0.776} & 0.196 / 0.332 & 0.654 / 1.307 & 0.328 / 0.654 & 0.298 / 0.602 & 0.408 / 0.734 & 10.75 / 17.45\\
\textbf{Joint View Vertically (Ours)} & 0.652 / 0.839 & \textbf{0.186} / \textbf{0.309} & \textbf{0.523} / \textbf{1.091} & 0.331 / 0.634 & \underline{0.267} / \underline{0.547} & 0.392 / \underline{0.684} & 10.92 / 17.70\\
AgentFormer \cite{yuan_agentformer_2021} & \textbf{0.482} / 0.794 & 0.237 / 0.456 & 0.622 / 1.310 & \underline{0.285} / \underline{0.564} & 0.296 / 0.624 & \underline{0.384} / 0.749 & 9.67 / 16.92\\
\textbf{Joint AgentFormer (Ours)} & \underline{0.485} / 0.798 & \underline{0.186} / \underline{0.320} & \underline{0.590} / \underline{1.219} & \textbf{0.271} / $\textbf{0.513}$ & \textbf{0.252} / \textbf{0.509} & \textbf{0.357} / \textbf{0.672} & \textbf{9.56} / 16.59\\

\hline
\end{tabular}
}  % end resizebox
}  % end subfloat
% \end{table*}

\subfloat[$CR_{JADE}$ / $CR_{mean}$]{
\label{table:all_CR_of_best-JADE_sample_mean_collision_rate}
\resizebox{\textwidth}{!}{
\begin{tabular}{c|ccccc|c||c}
\hline
\noalign{\smallskip}
 & \multicolumn{7}{c}{$\text{CR}_{mean} / \text{CR}_{JADE} \downarrow$  (m),  $K=20$ samples} \\
\cline{2-8}
\noalign{\smallskip} 
\;\;\;\;\;\;\;\; \text{Dataset \footnotesize(mean \# peds)} $\rightarrow$ & ETH \footnotesize(1.4) & HOTEL \footnotesize(2.7) & UNIV \footnotesize(25.7) & ZARA1 \footnotesize(3.3) & ZARA2 \footnotesize(5.9) & ETH / UCY Avg. & SDD TrajNet \footnotesize(1.5)\\
\noalign{\smallskip}
\hline
\noalign{\smallskip}
S-GAN \cite{gupta_social_2018} & 0.015 / \textbf{0.045} & 0.031 / \underline{0.090} & \underline{0.165} / \textbf{0.251} & 0.060 / 0.185 & 0.083 / \underline{0.195} & 0.071 / \underline{0.153} & 0.00 / 0.00\\
Trajectron++ \cite{salzmann_trajectron_2021} & 0.025 / 0.137 & 0.044 / 0.271 & 0.281 / 0.489 & 0.088 / 0.466 & 0.126 / 0.456 & 0.113 / 0.364 & 0.00 / 0.00\\
PECNet \cite{mangalam2020not} & 0.014 / 0.115 & 0.043 / 0.269 & 0.218 / 0.409 & 0.059 / 0.396 & 0.128 / 0.455 & 0.092 / 0.329 & 0.00 / 0.03\\
Y-Net \cite{mangalam_goals_2020} & 0.016 / 0.141 & 0.039 / 0.250 & 0.265 / 0.482 & 0.100 / 0.513 & 0.134 / 0.480 & 0.111 / 0.373 & 0.00 / 0.00\\
MemoNet \cite{Xu2022-ki} & 0.014 / 0.160 & 0.040 / 0.301 & 0.206 / 0.415 & 0.065 / 0.445 & 0.136 / 0.483 & 0.092 / 0.361 & 0.00 / 0.00\\
View Vertically \cite{Wong2022-kx} & 0.014 / 0.090 & 0.029 / 0.203 & 0.212 / 0.428 & 0.045 / 0.233 & 0.082 / 0.316 & 0.077 / 0.254 & 0.00 / 0.00\\
\textbf{Joint View Vertically (Ours)} & \textbf{0.011} / 0.076 & 0.026 / 0.168 & 0.276 / 0.484 & 0.045 / 0.262 & 0.081 / 0.349 & 0.088 / 0.268 & 0.00 / 0.00\\
AgentFormer \cite{yuan_agentformer_2021} & 0.016 / 0.068 & \underline{0.022} / \textbf{0.084} & 0.204 / 0.362 & \underline{0.021} / \textbf{0.088} & \textbf{0.054} / \textbf{0.139} & \underline{0.063} / \textbf{0.148} & \textbf{0.00} / \textbf{0.00}\\
\textbf{Joint AgentFormer (Ours)} & \underline{0.013} / \underline{0.064} & \textbf{0.019} / 0.094 & \textbf{0.163} / \underline{0.333} & \textbf{0.021} / \underline{0.100} & \underline{0.055} / 0.203 & \textbf{0.054} / 0.159 & \underline{0.00} / \underline{0.00}\\
\hline
Ground Truth & 0.000 & 0.001 & 0.021 & 0.000 & 0.002 & 0.005 & 0.00\\

\hline
\end{tabular}
}  % end resizebox
}  % end subfloat

\subfloat[$\min \textit{ADE} / \textit{FDE}$]{
\label{table:all_min_ADE_min_FDE}
\resizebox{\textwidth}{!}{
\begin{tabular}{c|ccccc|c||c}
\hline
\noalign{\smallskip}
 & \multicolumn{7}{c}{$\min \text{ADE}_{20} / \text{FDE}_{20} \downarrow$  (m),  $K=20$ samples} \\
\cline{2-8}
\noalign{\smallskip} 
\;\;\;\;\;\;\;\; \text{Dataset \footnotesize(mean \# peds)} $\rightarrow$ & ETH \footnotesize(1.4) & HOTEL \footnotesize(2.7) & UNIV \footnotesize(25.7) & ZARA1 \footnotesize(3.3) & ZARA2 \footnotesize(5.9) & ETH / UCY Avg. & SDD TrajNet \footnotesize(1.5)\\
\noalign{\smallskip}
\hline
\noalign{\smallskip}
S-GAN \cite{gupta_social_2018} & 0.876 / 1.656 & 0.461 / 0.920 & 0.639 / 1.343 & 0.379 / 0.816 & 0.285 / 0.600 & 0.528 / 1.067 & 12.74 / 22.65\\
Trajectron++ \cite{salzmann_trajectron_2021} & 0.669 / 1.183 & 0.185 / 0.283 & 0.303 / 0.541 & 0.249 / 0.414 & 0.175 / 0.319 & 0.316 / 0.548 & 10.18 / 15.76\\
PECNet \cite{mangalam2020not} & 0.562 / 0.985 & 0.192 / 0.332 & 0.336 / 0.630 & 0.243 / 0.468 & 0.179 / 0.345 & 0.302 / 0.552 & 9.34 / 16.10\\
Y-Net \cite{mangalam_goals_2020} & \textbf{0.398} / \textbf{0.571} & \underline{0.123} / 0.189 & 0.310 / 0.598 & 0.258 / 0.491 & 0.198 / 0.389 & 0.257 / 0.448 & 8.15 / \textbf{12.80}\\
MemoNet \cite{Xu2022-ki} & \underline{0.410} / \underline{0.636} & \textbf{0.113} / \textbf{0.173} & \textbf{0.244} / \textbf{0.433} & \underline{0.184} / \underline{0.320} & \underline{0.143} / 0.248 & \textbf{0.219} / \textbf{0.362} & \textbf{7.97} / \underline{12.82}\\
View Vertically \cite{Wong2022-kx} & 0.569 / 0.691 & 0.124 / \underline{0.188} & 0.290 / 0.499 & 0.202 / 0.356 & 0.147 / 0.257 & 0.267 / 0.398 & 9.34 / 14.67\\
\textbf{Joint View Vertically (Ours)} & 0.700 / 0.792 & 0.129 / 0.196 & 0.267 / 0.474 & 0.219 / 0.359 & 0.144 / 0.247 & 0.292 / 0.414 & 9.62 / 15.07\\
AgentFormer \cite{yuan_agentformer_2021} & 0.451 / 0.748 & 0.142 / 0.225 & \underline{0.254} / \underline{0.454} & \textbf{0.177} / \textbf{0.304} & \textbf{0.140} / \textbf{0.236} & \underline{0.233} / \underline{0.393} & \underline{8.01} / 13.24\\
\textbf{Joint AgentFormer (Ours)} & 0.473 / 0.792 & 0.135 / 0.212 & 0.285 / 0.505 & 0.189 / 0.321 & 0.144 / \underline{0.242} & 0.245 / 0.414 & 8.25 / 13.74\\

\hline
\end{tabular}
}  % end resizebox
}  % end subfloat

\end{table*}
% =============== END Main baseline tables ===================================

% \vspace{-2mm}\noindent\textbf{JXDE.}
\noindent\textbf{\JXDE{}.}
Joint metrics (Table \ref{table:all_min_JADE_min_JFDE}) perform about 2x worse across the board as compared to marginal metrics (Table \ref{table:all_min_ADE_min_FDE}). This means that while methods can achieve excellent predictions for individual agents across different prediction samples, they perform a lot worse at producing good predictions for all agents within a single prediction sample. This provides evidence that marginal metrics are overly optimistic estimations of trajectory forecasting performance.

Our Joint AgentFormer achieves the best \JADE{} and \JFDE{} of all methods: 7\% better in \JADE{} and 10\% better in \JFDE{} over AgentFormer, the next-best method in ETH / UCY. Our Joint View Vertically also achieves a 4\% boost in \JADE{} and a 7\% boost in \JFDE{} over vanilla View Vertically. The performance of all methods wrt \JXDE{} is summarized in Table \ref{table:all_min_JADE_min_JFDE}.
For both of our methods, there is a significantly large improvement in \JXDE{} performance (18\% for Joint AgentFormer and for 10\% for Joint View Vertically) in the \texttt{zara2} scene, an environment with plenty of interactions caused by medium-density sequences (about 6 pedestrians per 20-frame sequence). We hypothesize that our optimization method causes increased performance particularly on high-density sequences, in which there exist more interactions than in low-density sequences. Another example is in the \texttt{univ} scene, the densest scene, in which Joint View Vertically achieves a 20\% improvement, and Joint AgentFormer achieves a 5\% improvement. It is worthy to note that joint optimization results on a smaller improvement margin on scenes with few interactions (\texttt{eth}, \texttt{hotel}, and \texttt{SDD TrajNet}). Perhaps adding in consideration of joint dynamics in these simple scenes may confuse the model and impede accurate prediction.
% too many interactions that the causal structure of agent interaction behavior becomes too entangled, as in high-density sequences (\eg \textt{univ}. However, View Ver

While our method does not achieve the best \XDE{} (Table \ref{table:all_min_ADE_min_FDE}), we argue that the decrease in \XDE{} is worth the improvement in \JXDE{}. As seen in the trajectory visualizations in Figure \ref{fig:traj_viz}, which compares predictions from our method to that of View Vertically, a method may achieve excellent \XDE{} yet still fall short at producing natural and socially-compliant trajectories for certain agents.

\vspace{2mm}\noindent\textbf{Collision Rate.}
Our Joint AgentFormer performs best across the board with respect to collision rate, as seen in Table \ref{table:all_CR_of_best-JADE_sample_mean_collision_rate}. We gain an improvement in collision rate as compared with the baselines although we do not optimize explicitly with respect to it, substantiating our claim that optimizing for joint performance also leads naturally to a decrease in collision rate.  

\begin{figure}[t]
\begin{center}
  \includegraphics[width=.5\textwidth]{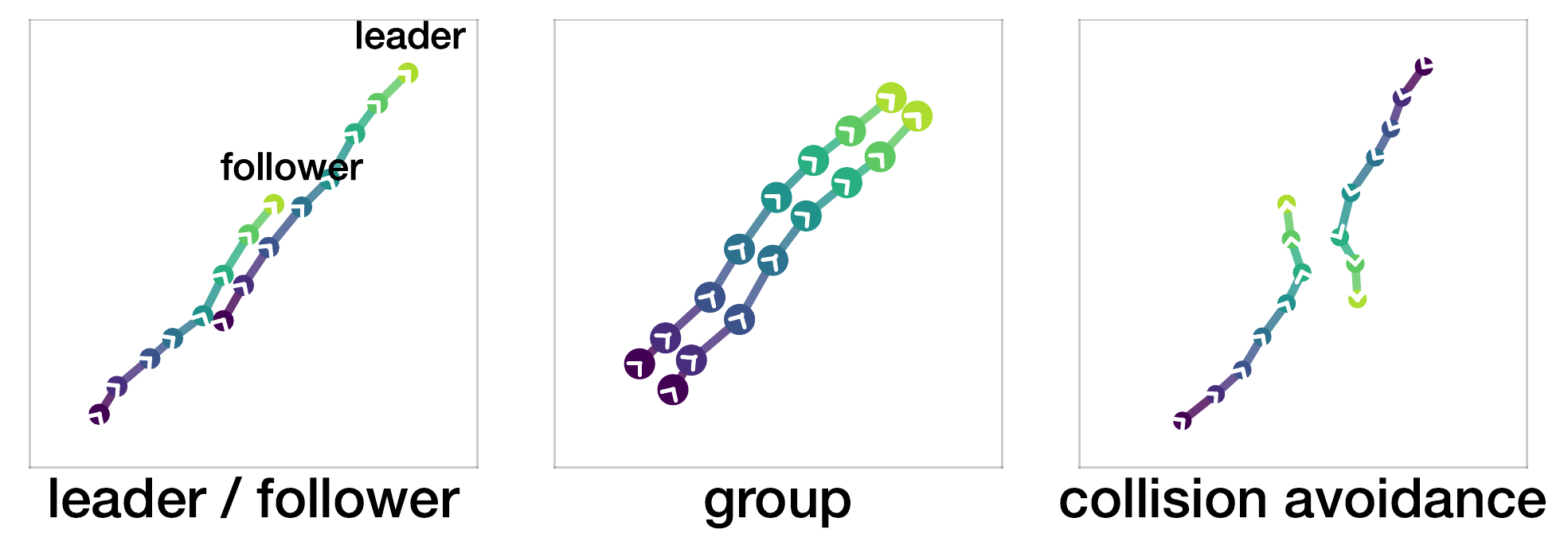}
\end{center}
\caption{Interaction Categories.}
\label{fig:int_cats}
\end{figure}

% ============ BEGIN Interaction Cats CR_mean AGG ============
\setlength{\tabcolsep}{2.1pt}
\begin{table}[t]
\caption{\footnotesize Comparison of collision rate performance of AgentFormer vs. Our Joint AgentFormer on different interaction categories in the ETH / UCY dataset.
In parentheses below each interaction category name shows the proportion of pedestrians belonging to that category across all 20-frame sequences, out of 34161 pedestrians total.
}
\label{table:int_cats}
% \resizebox{.99\columnwidth}{!}{%}
\begin{tabular*}{\columnwidth}{c|ccc|c}
\hline
% \noalign{\smallskip}
 & \multicolumn{4}{c}{$CR_{mean} \downarrow$ }\\
\cline{2-5}
\thead{Interaction Category \\ \scriptsize (proportion peds in category)} $\rightarrow$ & \thead{group\\ \\ \scriptsize (0.44)} & \thead{collison\\ avoidance\\ \scriptsize(0.61)} & \thead{leader-\\follower \\ \scriptsize (0.03)} & \thead{ ETH / UCY \\ aggregate\\ \scriptsize(1.0)} \\
\hline
\noalign{\smallskip}
AgentFormer \cite{yuan_agentformer_2021} & 0.103 & 0.201  &  0.124 & 0.063 \\
Joint AgentFormer (Ours) &  \textbf{0.084}   &  \textbf{0.180}  & \textbf{0.103}  & \textbf{0.053}  \\
\noalign{\smallskip}
\hline
\noalign{\smallskip}
ground-truth & 0.010 &  0.011  & 0.028 & 0.005 \\
\hline
\end{tabular*}%
% }% end resizebox
\end{table}
% ============ END Interaction Cats CR_mean AGG ============

\vspace{2mm}\noindent\textbf{Interaction Categories.}
We studied the effect of our improvements across different cross sections of the data. Specifically, we heuristically define 3 different categories of interactions present within the data:
% \textit{static}, \textit{linear}, \textit{non-linear},
\textit{group}, \textit{leader-follower}, and \textit{collision-avoidance}. 
% \textit{Static}, \textit{linear}, and \textit{non-linear} are single-agent categories; they are defined upon individual agents independently of other agents, and are disjoint categories. On the other hand, 
% \textit{Group}, \textit{leader-follower}, and \textit{collision-avoidance} are 
These interaction categories are used to
% ; they are also 
categorize individual agents based on whether that agent interacts in a certain way with at least one other agent within a 20-frame sequence. The categories are not disjoint, so each agent may fall into none, one, or multiple categories.
\textit{Group} defines an agent who moves in parallel with another agent; \textit{leader-follower} defines an agent who either is moving behind or before another agent in the same direction as that other agent; \textit{collision-avoidance} defines an agent that comes within a distance threshold to another agent that is not moving in the same direction. Definitions of the heuristics used to create these categories can be found in the supplementary material; we take inspiration from and modify the heuristics used in~\cite{Kothari2022-hl}. Figure \ref{fig:int_cats} shows examples of pedestrians within each category. 

Using our defined categories, we examine collision rate performance on the 
% \texttt{univ} scene of the 
ETH / UCY dataset. 
% As readers can see in Table \ref{table:all_CR_of_best-JADE_sample_mean_collision_rate}, the performance on \texttt{univ} is particularly poor with respect to collision rate, likely due to its high pedestrian density of 25.7 pedestrians per 20-frame sequence, and thus large proportion of socially-interacting trajectories.
We highlight the success of our method with respect to modeling 
% especially in dense scenes, evidenced by our method achieving a sizable improvement with respect to collision rate. In particular, we note that our method achieves an especially great improvement with respect to 
of pedestrians involved in interactions, 
% within the \texttt{univ} scene, 
as seen in Table \ref{table:int_cats}: a 23\% decrease in \textit{group}, a 19\% decrease in \textit{collision-avoidance}, and a 24\% decrease in \textit{leader-follower}. These results substantiate our claim that our method improves social compliance.

% ============ BEGIN Interaction Cats CR_mean UNIV ============
% \setlength{\tabcolsep}{2.1pt}
% \begin{table}[t]
% \caption{Comparison of collision rate performance of AgentFormer vs. Our method on different interaction categories in the \texttt{univ} scene.
% In parentheses below each interaction category name shows the proportion of pedestrians belonging to that category across all $20$-frame sequences in the \texttt{univ} scene, out of $24334$ pedestrians total.
% }
% \label{table:int_cats}
% \resizebox{.99\columnwidth}{!}{
% \begin{tabular*}{\columnwidth}{c|ccc|c}
% \hline
% \noalign{\smallskip}
%  & \multicolumn{4}{c}{\text{ {CR}_{mean} \downarrow} }\\
% \cline{2-5}
% \thead{Interaction Category \\ \scriptsize (proportion peds in category)} \rightarrow & \thead{group\\ \\ \scriptsize (0.41)} & \thead{collison\\ avoidance\\ \scriptsize(0.76)} & \thead{leader-\\follower \\ \scriptsize (0.03)} & \thead{\texttt{univ}\\ aggregate\\ \scriptsize(1.0)} \\
% \hline
% \noalign{\smallskip}
% AgentFormer \cite{yuan_agentformer_2021} & 0.25 & 0.33   &  0.33  & 0.20 \\
% Joint AgentFormer (Ours) &  \textbf{0.20}   &  \textbf{0.27}  & \textbf{0.25}  & \textbf{0.16}  \\
% \noalign{\smallskip}
% \hline
% \noalign{\smallskip}
% ground-truth & 0.04 & 0.04 & 0.14 & 0.02\\
% \hline
% \end{tabular*}
% } % end resizebox
% \end{table}
% ============ END Interaction Cats CR_mean UNIV ============

% ===================== Ablation Tables BEGIN ========================
\begin{table}[t]
\caption{\small Ablation studies on the usage of Marginal and Joint loss terms in the two training steps of AgentFormer optimization.}
\label{table:ablations}
\resizebox{.98\columnwidth}{!}{
\subfloat[Ablation Results for \textit{JADE} / \textit{JFDE}.]
{
\label{table:ablation_JXDE}
\begin{tabular}{c|cc|cc|c}
\hline
\noalign{\smallskip}
\multicolumn{6}{c}{$\min \text{JADE}_{20} / \text{JFDE}_{20} \;\;\downarrow$ (meters)} \\
\noalign{\smallskip}
\hline
\noalign{\smallskip}
& \multicolumn{2}{c|}{training step 1} & \multicolumn{2}{c|}{training step 2} & \\
\noalign{\smallskip}
\cline{2-5}
\noalign{\smallskip} 
& \thead{Marginal\\ term} & \thead{Joint\\ term} & \thead{Marginal\\ term} & \thead{Joint\\ term} & 
\thead{ETH / UCY \\ Average}\\
\noalign{\smallskip}
\hline
\noalign{\smallskip}
AgentFormer \cite{yuan_agentformer_2021} & $\checkmark$ &  & $\checkmark$ & & 0.384 / 0.749\\
- & $\checkmark$ & & & $\checkmark$ &  0.365 / 0.694\\
- & &  $\checkmark$ & & $\checkmark$ & 0.386 / 0.734\\
% $\checkmark$ & &$\checkmark$  &   & 0.37 / 0.70\\
- &  $\checkmark$ & $\checkmark$  & & $\checkmark$ & \underline{0.358} / \underline{0.672}\\
Ours & $\checkmark$ & $\checkmark$  & $\checkmark$ & $\checkmark$  & \textbf{0.357} / \textbf{0.672}\\
\hline
\end{tabular}
} % end resize box
}\\
\hfill
% ===================== CR ========================
\resizebox{.98\columnwidth}{!}{
\subfloat[Ablation Results for mean Collison Rate.]{
\label{table:ablation_CR}
\begin{tabular}{c|cc|cc|c}
\hline
\noalign{\smallskip}
\multicolumn{6}{c}{$\text{CR}_{mean} \;\;\downarrow$; $K=20$ samples }\\
\noalign{\smallskip}
\hline
\noalign{\smallskip}
& \multicolumn{2}{c|}{training step 1} & \multicolumn{2}{c|}{training step 2} & \\
\noalign{\smallskip}
\cline{2-5}
\noalign{\smallskip} 
& \thead{Marginal\\ term} & \thead{Joint\\ term} & \thead{Marginal\\ term} & \thead{Joint\\ term} & 
\thead{ETH / UCY \\ Average}
\\
\noalign{\smallskip}
\hline
\noalign{\smallskip}
AgentFormer \cite{yuan_agentformer_2021} & $\checkmark$ &  & $\checkmark$ & & 0.063\\
- & $\checkmark$ & & & $\checkmark$ &  0.064\\
- & &  $\checkmark$ & & $\checkmark$ &  \textbf{0.051}\\
% $\checkmark$ & &$\checkmark$  &   & \underline{0.008} / \underline{0.052}\\
- &  $\checkmark$ & $\checkmark$  & & $\checkmark$ &  \underline{0.053}\\
Ours &  $\checkmark$ & $\checkmark$  & $\checkmark$ & $\checkmark$ & 0.054 \\
\hline
\end{tabular}
} % end subfloat
} % end resize box

% ===================== ADE / FDE ========================
\resizebox{.98\columnwidth}{!}{
\subfloat[Ablation Results for \ADE{} / \textit{FDE}.]{
\label{table:ablation_XDE_supp}
\begin{tabular}{c|cc|cc|c}
\hline
\noalign{\smallskip}
\multicolumn{6}{c}{$\min \ADE{}_{20} / \FDE{}_{20} \;\;\downarrow$; $K=20$ samples }\\
\noalign{\smallskip}
\hline
\noalign{\smallskip}
& \multicolumn{2}{c|}{training step 1} & \multicolumn{2}{c|}{training step 2} & \\
\noalign{\smallskip}
\cline{2-5}
\noalign{\smallskip} 
& \thead{Marginal\\ term} & \thead{Joint\\ term} & \thead{Marginal\\ term} & \thead{Joint\\ term} & 
\thead{ETH / UCY \\ Average}
\\
\noalign{\smallskip}
\hline
\noalign{\smallskip}
AgentFormer \cite{yuan_agentformer_2021} & $\checkmark$ &  & $\checkmark$ & & \textbf{0.233} / \textbf{0.393}\\
- & $\checkmark$ & & & $\checkmark$ & 0.255 / 0.436 \\
- & &  $\checkmark$ & & $\checkmark$ &  0.293 / 0.516\\
% $\checkmark$ & &$\checkmark$  &   & \underline{0.008} / \underline{0.052}\\
- &  $\checkmark$ & $\checkmark$  & & $\checkmark$ &  0.258 / 0.439 \\
Ours &  $\checkmark$ & $\checkmark$  & $\checkmark$ & $\checkmark$ & \underline{0.245} / \underline{0.414}\\
\hline
\end{tabular}
} % end subfloat
} % end resize box
\end{table}
% ===================== Ablation tables END ========================

\vspace{2mm}\noindent\textbf{Ablation Studies.}
We perform ablations to study the effect on joint metric performance of using the marginal reconstruction loss and / or the joint reconstruction loss in both steps of AgentFormer training, as seen in Table \ref{table:ablations}.

An interesting observation is that training AgentFormer with only joint loss during both training steps (line 3 of Table \ref{table:ablation_JXDE}) does not result in as good \JXDE{} as compared with training with both marginal and joint loss in the first step, and then only joint loss in the second (as we do in Our Joint AgentFormer, line 4 of Table \ref{table:ablation_JXDE}). A possible reason for this is because the joint prediction problem is inherently more difficult than the marginal prediction problem, due to having to optimize for joint performance of multiple agents rather than individual agents independently. As the joint loss function more naturally captures social compliance for the joint prediction problem than the marginal loss function, as previously established, optimizing it acquires the difficulty of the joint prediction problem. \erica{I have a feeling Kris is going to say this is not clear / false reasoning. Perhaps you can confirm? I can also just leave out my hypothesis since I don't test it anyways. My point I want to make is that I think marginal loss space is smoother than the joint loss space and thus easier to optimize, and I guess this since the marginal prediction is an easier problem, since you don't have to consider how backpropping will affect joint performance when optimizing. but optimizing wrt marginal metrics naturally lends itself to \textit{some} goodness in joint metrics, since the two after all are related, so you see joint earlier / more easilty in optimizing. On the other hand, joint optimization may be more ``finnicky" to optimize because the loss space is less smooth. So my guess is that is why the third row does not do as well as the last row. But after all this is an untested hypothesis.}
 
Another point of interest with regard to the two-step training procedure of AgentFormer is that models trained with joint loss during the first AgentFormer training step show greater improvement in mean collision rate (Line 3 and 4 in Table \ref{table:ablation_CR}). This may be due to the fact that the first training step accounts for the majority of AgentFormer training, as that is when most weights of the CVAE are trained; the second training step only learns the weights of the \textit{Trajectory Sampler}, which account for only a small fraction of the entire network.

\section{Conclusion}
% \vspace{-2mm}
In this paper, we illustrated the importance of joint metrics in addition to marginal metrics, presenting a comprehensive benchmark evaluation of SOTA methods in multi-agent trajectory forecasting with respect to \JXDE{} and collision rate. Further, we proposed a new approach that optimizes for joint performance rather than only marginal performance. Our Joint AgentFormer achieves state-of-the-art results with respect to \JXDE{}, outperforming all previous methods. Our Joint View Vertically also outperforms the original View Vertically. We also highlight the improvements achieved in collision rate, showing that joint optimization may also lead to improved interaction modeling. Finally, we push for the widespread adoption of joint metrics in trajectory forecasting evaluation, so we can achieve more realistic estimation of method performance; as well as the adoption of joint loss functions in optimization, so we can achieve better interaction modeling of multi-modal futures.

\vspace{4mm}
\noindent\textbf{Acknowledgements.} This work is supported in part by the Ford Foundation Fellowship. Thank you also to Ye Yuan and Aaditya Singh for their feedback on the manuscript.

{\small
\bibliographystyle{ieee_fullname}
\bibliography{ye_hana_erica, paperpile}

\begin{thebibliography}{10}\itemsep=-1pt

\bibitem{alahi_social_2016}
Alexandre Alahi, Kratarth Goel, Vignesh Ramanathan, Alexandre Robicquet, Li
  Fei-Fei, and Silvio Savarese.
\newblock Social {LSTM}: {Human} {Trajectory} {Prediction} in {Crowded}
  {Spaces}.
\newblock In {\em 2016 {IEEE} {Conference} on {Computer} {Vision} and {Pattern}
  {Recognition} ({CVPR})}, pages 961--971, Las Vegas, NV, USA, June 2016. IEEE.

\bibitem{alet_neural_2019}
Ferran Alet, Erica Weng, Tomás Lozano-Pérez, and Leslie~Pack Kaelbling.
\newblock Neural {Relational} {Inference} with {Fast} {Modular}
  {Meta}-learning.
\newblock page~12.

\bibitem{Amirloo2022-se}
Elmira Amirloo, Amir Rasouli, Peter Lakner, Mohsen Rohani, and Jun Luo.
\newblock {LatentFormer}: {Multi-Agent} {Transformer-Based} interaction
  modeling and trajectory prediction.
\newblock Mar. 2022.

\bibitem{bahari_svg-net_2021}
Mohammadhossein Bahari, Vahid Zehtab, Sadegh Khorasani, Sana Ayromlou, Saeed
  Saadatnejad, and Alexandre Alahi.
\newblock {SVG}-{Net}: {An} {SVG}-based {Trajectory} {Prediction} {Model}.
\newblock {\em arXiv:2110.03706 [cs]}, Oct. 2021.
\newblock arXiv: 2110.03706.

\bibitem{Cao2022-hq}
Yulong Cao, Chaowei Xiao, Anima Anandkumar, Danfei Xu, and Marco Pavone.
\newblock {AdvDO}: Realistic adversarial attacks for trajectory prediction.
\newblock In {\em Computer Vision -- {ECCV} 2022}, pages 36--52. Springer
  Nature Switzerland, 2022.

\bibitem{Cao2022-uv}
Zhangjie Cao, Erdem B{\i}y{\i}k, Guy Rosman, and Dorsa Sadigh.
\newblock Leveraging smooth attention prior for {Multi-Agent} trajectory
  prediction.
\newblock Mar. 2022.

\bibitem{chai_multipath_2019}
Yuning Chai, Benjamin Sapp, Mayank Bansal, and Dragomir Anguelov.
\newblock {MultiPath}: {Multiple} {Probabilistic} {Anchor} {Trajectory}
  {Hypotheses} for {Behavior} {Prediction}.
\newblock {\em arXiv:1910.05449 [cs, stat]}, Oct. 2019.
\newblock arXiv: 1910.05449.

\bibitem{chai2020multipath}
Yuning Chai, Benjamin Sapp, Mayank Bansal, and Dragomir Anguelov.
\newblock Multipath: Multiple probabilistic anchor trajectory hypotheses for
  behavior prediction.
\newblock In {\em Conference on Robot Learning}, pages 86--99. PMLR, 2020.

\bibitem{chang_argoverse_2019}
Ming-Fang Chang, Deva Ramanan, James Hays, John Lambert, Patsorn Sangkloy,
  Jagjeet Singh, Slawomir Bak, Andrew Hartnett, De Wang, Peter Carr, and Simon
  Lucey.
\newblock Argoverse: {3D} {Tracking} and {Forecasting} {With} {Rich} {Maps}.
\newblock In {\em 2019 {IEEE}/{CVF} {Conference} on {Computer} {Vision} and
  {Pattern} {Recognition} ({CVPR})}, pages 8740--8749, Long Beach, CA, USA,
  June 2019. IEEE.

\bibitem{chen_relational_2020}
Changan Chen, Sha Hu, Payam Nikdel, Greg Mori, and Manolis Savva.
\newblock Relational {Graph} {Learning} for {Crowd} {Navigation}.
\newblock {\em arXiv:1909.13165 [cs]}, Aug. 2020.
\newblock arXiv: 1909.13165.

\bibitem{Chiara2022-xa}
Luigi~Filippo Chiara, Pasquale Coscia, Sourav Das, Simone Calderara, Rita
  Cucchiara, and Lamberto Ballan.
\newblock Goal-driven self-attentive recurrent networks for trajectory
  prediction.
\newblock pages 2518--2527, Apr. 2022.

\bibitem{Ettinger2021-yl}
Scott Ettinger, Shuyang Cheng, Benjamin Caine, Chenxi Liu, Hang Zhao, Sabeek
  Pradhan, Yuning Chai, Ben Sapp, Charles Qi, Yin Zhou, Zoey Yang, Aurelien
  Chouard, Pei Sun, Jiquan Ngiam, Vijay Vasudevan, Alexander McCauley, Jonathon
  Shlens, and Dragomir Anguelov.
\newblock Large scale interactive motion forecasting for autonomous driving :
  The waymo open motion dataset.
\newblock Apr. 2021.

\bibitem{goodfellow2014generative}
Ian~J Goodfellow, Jean Pouget-Abadie, Mehdi Mirza, Bing Xu, David Warde-Farley,
  Sherjil Ozair, Aaron Courville, and Yoshua Bengio.
\newblock Generative adversarial networks.
\newblock {\em arXiv preprint arXiv:1406.2661}, 2014.

\bibitem{guan2020generative}
Jiaqi Guan, Ye Yuan, Kris~M Kitani, and Nicholas Rhinehart.
\newblock Generative hybrid representations for activity forecasting with
  no-regret learning.
\newblock In {\em Proceedings of the IEEE/CVF Conference on Computer Vision and
  Pattern Recognition}, pages 173--182, 2020.

\bibitem{gupta2018social}
Agrim Gupta, Justin Johnson, Li Fei-Fei, Silvio Savarese, and Alexandre Alahi.
\newblock Social gan: Socially acceptable trajectories with generative
  adversarial networks.
\newblock In {\em Proceedings of the IEEE Conference on Computer Vision and
  Pattern Recognition}, pages 2255--2264, 2018.

\bibitem{gupta_social_2018}
Agrim Gupta, Justin Johnson, Li Fei-Fei, Silvio Savarese, and Alexandre Alahi.
\newblock Social {GAN}: {Socially} {Acceptable} {Trajectories} {With}
  {Generative} {Adversarial} {Networks}.
\newblock pages 2255--2264, 2018.

\bibitem{helbing_social_1995}
Dirk Helbing and Peter Molnar.
\newblock Social {Force} {Model} for {Pedestrian} {Dynamics}.
\newblock {\em Physical Review E}, 51(5):4282--4286, May 1995.
\newblock arXiv: cond-mat/9805244.

\bibitem{Huang2021-sv}
Zhiyu Huang, Xiaoyu Mo, and Chen Lv.
\newblock Multi-modal motion prediction with transformer-based neural network
  for autonomous driving.
\newblock Sept. 2021.

\bibitem{ivanovic2019trajectron}
Boris Ivanovic and Marco Pavone.
\newblock The trajectron: Probabilistic multi-agent trajectory modeling with
  dynamic spatiotemporal graphs.
\newblock In {\em Proceedings of the IEEE/CVF International Conference on
  Computer Vision}, pages 2375--2384, 2019.

\bibitem{kingma2013auto}
Diederik~P Kingma and Max Welling.
\newblock Auto-encoding variational bayes.
\newblock {\em arXiv preprint arXiv:1312.6114}, 2013.

\bibitem{kipf2018neural}
Thomas Kipf, Ethan Fetaya, Kuan-Chieh Wang, Max Welling, and Richard Zemel.
\newblock Neural relational inference for interacting systems.
\newblock In {\em International Conference on Machine Learning}, pages
  2688--2697. PMLR, 2018.

\bibitem{kosaraju2019social}
Vineet Kosaraju, Amir Sadeghian, Roberto Mart{\'\i}n-Mart{\'\i}n, Ian~D Reid,
  Hamid Rezatofighi, and Silvio Savarese.
\newblock Social-bigat: multimodal trajectory forecasting using bicycle-gan and
  graph attention networks.
\newblock In {\em Advances in Neural Information Processing Systems 2019}.
  Neural Information Processing Systems (NIPS), 2019.

\bibitem{Kothari2022-cy}
Parth Kothari and Alexandre Alahi.
\newblock Safety-compliant generative adversarial networks for human trajectory
  forecasting.
\newblock Sept. 2022.

\bibitem{kothari_human_2021}
Parth Kothari, Sven Kreiss, and Alexandre Alahi.
\newblock Human {Trajectory} {Forecasting} in {Crowds}: {A} {Deep} {Learning}
  {Perspective}.
\newblock {\em arXiv:2007.03639 [cs]}, Jan. 2021.
\newblock arXiv: 2007.03639.

\bibitem{Kothari2022-hl}
Parth Kothari, Sven Kreiss, and Alexandre Alahi.
\newblock Human trajectory forecasting in crowds: A deep learning perspective.
\newblock {\em IEEE Trans. Intell. Transp. Syst.}, 23(7):7386--7400, July 2022.

\bibitem{kothari_interpretable_2021}
Parth Kothari, Brian Sifringer, and Alexandre Alahi.
\newblock Interpretable {Social} {Anchors} for {Human} {Trajectory}
  {Forecasting} in {Crowds}.
\newblock {\em arXiv:2105.03136 [cs]}, May 2021.
\newblock arXiv: 2105.03136.

\bibitem{lee2017desire}
Namhoon Lee, Wongun Choi, Paul Vernaza, Christopher~B Choy, Philip~HS Torr, and
  Manmohan Chandraker.
\newblock Desire: Distant future prediction in dynamic scenes with interacting
  agents.
\newblock In {\em Proceedings of the IEEE Conference on Computer Vision and
  Pattern Recognition}, pages 336--345, 2017.

\bibitem{lerner2007crowds}
Alon Lerner, Yiorgos Chrysanthou, and Dani Lischinski.
\newblock Crowds by example.
\newblock In {\em Computer graphics forum}, volume~26, pages 655--664. Wiley
  Online Library, 2007.

\bibitem{lerner_crowds_2007}
Alon Lerner, Yiorgos Chrysanthou, and Dani Lischinski.
\newblock Crowds by {Example}.
\newblock {\em Computer Graphics Forum}, 26(3):655--664, 2007.
\newblock \_eprint:
  https://onlinelibrary.wiley.com/doi/pdf/10.1111/j.1467-8659.2007.01089.x.

\bibitem{liang_garden_2020}
Junwei Liang, Lu Jiang, Kevin Murphy, Ting Yu, and Alexander Hauptmann.
\newblock The {Garden} of {Forking} {Paths}: {Towards} {Multi}-{Future}
  {Trajectory} {Prediction}.
\newblock {\em arXiv:1912.06445 [cs]}, Mar. 2020.
\newblock arXiv: 1912.06445.

\bibitem{Liang2020-of}
Ming Liang, Bin Yang, Rui Hu, Yun Chen, Renjie Liao, Song Feng, and Raquel
  Urtasun.
\newblock Learning lane graph representations for motion forecasting.
\newblock July 2020.

\bibitem{liu_energy-based_2021}
Minghuan Liu, Tairan He, Minkai Xu, and Weinan Zhang.
\newblock Energy-{Based} {Imitation} {Learning}.
\newblock {\em arXiv:2004.09395 [cs, stat]}, Apr. 2021.
\newblock arXiv: 2004.09395.

\bibitem{liu_towards_2021}
Yuejiang Liu, Riccardo Cadei, Jonas Schweizer, Sherwin Bahmani, and Alexandre
  Alahi.
\newblock Towards {Robust} and {Adaptive} {Motion} {Forecasting}: {A} {Causal}
  {Representation} {Perspective}.
\newblock {\em arXiv:2111.14820 [cs]}, Nov. 2021.
\newblock arXiv: 2111.14820.

\bibitem{liu_social_2021}
Yuejiang Liu, Qi Yan, and Alexandre Alahi.
\newblock Social {NCE}: {Contrastive} {Learning} of {Socially}-aware {Motion}
  {Representations}.
\newblock {\em arXiv:2012.11717 [cs]}, Aug. 2021.
\newblock arXiv: 2012.11717.

\bibitem{mangalam_goals_2020}
Karttikeya Mangalam, Yang An, Harshayu Girase, and Jitendra Malik.
\newblock From {Goals}, {Waypoints} \& {Paths} {To} {Long} {Term} {Human}
  {Trajectory} {Forecasting}.
\newblock {\em arXiv:2012.01526 [cs]}, Dec. 2020.
\newblock arXiv: 2012.01526.

\bibitem{mangalam_it_2020}
Karttikeya Mangalam, Harshayu Girase, Shreyas Agarwal, Kuan-Hui Lee, Ehsan
  Adeli, Jitendra Malik, and Adrien Gaidon.
\newblock It {Is} {Not} the {Journey} but the {Destination}: {Endpoint}
  {Conditioned} {Trajectory} {Prediction}.
\newblock {\em arXiv:2004.02025 [cs]}, July 2020.
\newblock arXiv: 2004.02025.

\bibitem{mangalam2020not}
Karttikeya Mangalam, Harshayu Girase, Shreyas Agarwal, Kuan-Hui Lee, Ehsan
  Adeli, Jitendra Malik, and Adrien Gaidon.
\newblock It is not the journey but the destination: Endpoint conditioned
  trajectory prediction.
\newblock In {\em European Conference on Computer Vision}, pages 759--776.
  Springer, 2020.

\bibitem{mohamed_social-stgcnn_2020}
Abduallah Mohamed, Kun Qian, Mohamed Elhoseiny, and Christian Claudel.
\newblock Social-{STGCNN}: {A} {Social} {Spatio}-{Temporal} {Graph}
  {Convolutional} {Neural} {Network} for {Human} {Trajectory} {Prediction}.
\newblock {\em arXiv:2002.11927 [cs]}, Mar. 2020.
\newblock arXiv: 2002.11927 version: 3.

\bibitem{pellegrini_youll_2009}
S. Pellegrini, A. Ess, K. Schindler, and L. van Gool.
\newblock You'll never walk alone: {Modeling} social behavior for multi-target
  tracking.
\newblock In {\em 2009 {IEEE} 12th {International} {Conference} on {Computer}
  {Vision}}, pages 261--268, Sept. 2009.
\newblock ISSN: 2380-7504.

\bibitem{rezende2015variational}
Danilo Rezende and Shakir Mohamed.
\newblock Variational inference with normalizing flows.
\newblock In {\em International Conference on Machine Learning}, pages
  1530--1538. PMLR, 2015.

\bibitem{rhinehart2018r2p2}
Nicholas Rhinehart, Kris~M Kitani, and Paul Vernaza.
\newblock R2p2: A reparameterized pushforward policy for diverse, precise
  generative path forecasting.
\newblock In {\em Proceedings of the European Conference on Computer Vision
  (ECCV)}, pages 772--788, 2018.

\bibitem{rhinehart2019precog}
Nicholas Rhinehart, Rowan McAllister, Kris Kitani, and Sergey Levine.
\newblock Precog: Prediction conditioned on goals in visual multi-agent
  settings.
\newblock In {\em Proceedings of the IEEE/CVF International Conference on
  Computer Vision}, pages 2821--2830, 2019.

\bibitem{robicquet_learning_2016}
Alexandre Robicquet, Amir Sadeghian, Alexandre Alahi, and S. Savarese.
\newblock Learning {Social} {Etiquette}: {Human} {Trajectory} {Understanding}
  {In} {Crowded} {Scenes}.
\newblock In {\em {ECCV}}, 2016.

\bibitem{saadatnejad_are_2021}
Saeed Saadatnejad, Mohammadhossein Bahari, Pedram Khorsandi, Mohammad Saneian,
  Seyed-Mohsen Moosavi-Dezfooli, and Alexandre Alahi.
\newblock Are socially-aware trajectory prediction models really
  socially-aware?
\newblock {\em arXiv:2108.10879 [cs]}, Aug. 2021.
\newblock arXiv: 2108.10879.

\bibitem{sadeghian2019sophie}
Amir Sadeghian, Vineet Kosaraju, Ali Sadeghian, Noriaki Hirose, Hamid
  Rezatofighi, and Silvio Savarese.
\newblock Sophie: An attentive gan for predicting paths compliant to social and
  physical constraints.
\newblock In {\em Proceedings of the IEEE/CVF Conference on Computer Vision and
  Pattern Recognition}, pages 1349--1358, 2019.

\bibitem{salzmann_trajectron_2021}
Tim Salzmann, Boris Ivanovic, Punarjay Chakravarty, and Marco Pavone.
\newblock Trajectron++: {Dynamically}-{Feasible} {Trajectory} {Forecasting}
  {With} {Heterogeneous} {Data}.
\newblock {\em arXiv:2001.03093 [cs]}, Jan. 2021.
\newblock arXiv: 2001.03093.

\bibitem{sohn_a2x_2021}
Samuel~S. Sohn, Mihee Lee, Seonghyeon Moon, Gang Qiao, Muhammad Usman, Sejong
  Yoon, Vladimir Pavlovic, and Mubbasir Kapadia.
\newblock {A2X}: {An} {Agent} and {Environment} {Interaction} {Benchmark} for
  {Multimodal} {Human} {Trajectory} {Prediction}.
\newblock In {\em Motion, {Interaction} and {Games}}, pages 1--9, Virtual Event
  Switzerland, Nov. 2021. ACM.

\bibitem{su_trajectory_2022}
Yuchao Su, Jie Du, Yuanman Li, Xia Li, Rongqin Liang, Zhongyun Hua, and Jiantao
  Zhou.
\newblock Trajectory {Forecasting} {Based} on {Prior}-{Aware} {Directed}
  {Graph} {Convolutional} {Neural} {Network}.
\newblock {\em IEEE Transactions on Intelligent Transportation Systems}, pages
  1--13, 2022.
\newblock Conference Name: IEEE Transactions on Intelligent Transportation
  Systems.

\bibitem{sun_scalability_2020}
Pei Sun, Henrik Kretzschmar, Xerxes Dotiwalla, Aurelien Chouard, Vijaysai
  Patnaik, Paul Tsui, James Guo, Yin Zhou, Yuning Chai, Benjamin Caine, Vijay
  Vasudevan, Wei Han, Jiquan Ngiam, Hang Zhao, Aleksei Timofeev, Scott
  Ettinger, Maxim Krivokon, Amy Gao, Aditya Joshi, Yu Zhang, Jonathon Shlens,
  Zhifeng Chen, and Dragomir Anguelov.
\newblock Scalability in {Perception} for {Autonomous} {Driving}: {Waymo}
  {Open} {Dataset}.
\newblock In {\em 2020 {IEEE}/{CVF} {Conference} on {Computer} {Vision} and
  {Pattern} {Recognition} ({CVPR})}, pages 2443--2451, Seattle, WA, USA, June
  2020. IEEE.

\bibitem{Sun2022-mj}
Qiao Sun, Xin Huang, Junru Gu, Brian~C Williams, and Hang Zhao.
\newblock {M2I}: From factored marginal trajectory prediction to interactive
  prediction.
\newblock Feb. 2022.

\bibitem{tang2019multiple}
Yichuan~Charlie Tang and Ruslan Salakhutdinov.
\newblock Multiple futures prediction.
\newblock {\em arXiv preprint arXiv:1911.00997}, 2019.

\bibitem{weng2020joint}
Xinshuo Weng, Ye Yuan, and Kris Kitani.
\newblock Joint 3d tracking and forecasting with graph neural network and
  diversity sampling.
\newblock {\em arXiv preprint arXiv:2003.07847}, 2020.

\bibitem{Wong2022-kx}
Conghao Wong, Beihao Xia, Ziming Hong, Qinmu Peng, Wei Yuan, Qiong Cao, Yibo
  Yang, and Xinge You.
\newblock View vertically: A hierarchical network for trajectory prediction via
  fourier spectrums.
\newblock In {\em Computer Vision -- {ECCV} 2022}, pages 682--700. Springer
  Nature Switzerland, 2022.

\bibitem{Xu2022-ki}
Chenxin Xu, Weibo Mao, Wenjun Zhang, and Siheng Chen.
\newblock Remember intentions: {Retrospective-Memory-based} trajectory
  prediction.
\newblock Mar. 2022.

\bibitem{Yao2021-xn}
Yu Yao, Ella Atkins, Matthew Johnson-Roberson, Ram Vasudevan, and Xiaoxiao Du.
\newblock {BiTraP}: {Bi-Directional} pedestrian trajectory prediction with
  {Multi-Modal} goal estimation.
\newblock {\em IEEE Robotics and Automation Letters}, 6(2):1463--1470, Apr.
  2021.

\bibitem{Yu2020-df}
Cunjun Yu, Xiao Ma, Jiawei Ren, Haiyu Zhao, and Shuai Yi.
\newblock {Spatio-Temporal} graph transformer networks for pedestrian
  trajectory prediction.
\newblock In {\em Computer Vision -- {ECCV} 2020}, pages 507--523. Springer
  International Publishing, 2020.

\bibitem{yuan2019diverse}
Ye Yuan and Kris Kitani.
\newblock Diverse trajectory forecasting with determinantal point processes.
\newblock {\em arXiv preprint arXiv:1907.04967}, 2019.

\bibitem{Yuan2020-vw}
Ye Yuan and Kris Kitani.
\newblock {DLow}: Diversifying latent flows for diverse human motion
  prediction.
\newblock Mar. 2020.

\bibitem{yuan2020dlow}
Ye Yuan and Kris Kitani.
\newblock Dlow: Diversifying latent flows for diverse human motion prediction.
\newblock In {\em European Conference on Computer Vision}, pages 346--364.
  Springer, 2020.

\bibitem{Yuan2021-tp}
Ye Yuan, Xinshuo Weng, Yanglan Ou, and Kris Kitani.
\newblock {AgentFormer}: {Agent-Aware} transformers for {Socio-Temporal}
  {Multi-Agent} forecasting.
\newblock pages 9813--9823, Mar. 2021.

\bibitem{yuan_agentformer_2021}
Ye Yuan, Xinshuo Weng, Yanglan Ou, and Kris Kitani.
\newblock {AgentFormer}: {Agent}-{Aware} {Transformers} for {Socio}-{Temporal}
  {Multi}-{Agent} {Forecasting}.
\newblock {\em arXiv:2103.14023 [cs]}, Oct. 2021.
\newblock arXiv: 2103.14023.

\bibitem{Yue2022-os}
Jiangbei Yue, Dinesh Manocha, and He Wang.
\newblock Human trajectory prediction via neural social physics.
\newblock July 2022.

\bibitem{Zhang2022-dp}
Qingzhao Zhang, Shengtuo Hu, Jiachen Sun, Qi~Alfred Chen, and Z~Morley Mao.
\newblock On adversarial robustness of trajectory prediction for autonomous
  vehicles.
\newblock pages 15159--15168, Jan. 2022.

\bibitem{zhao_tnt_2020}
Hang Zhao, Jiyang Gao, Tian Lan, Chen Sun, Benjamin Sapp, Balakrishnan
  Varadarajan, Yue Shen, Yi Shen, Yuning Chai, Cordelia Schmid, Congcong Li,
  and Dragomir Anguelov.
\newblock {TNT}: {Target}-{driveN} {Trajectory} {Prediction}.
\newblock {\em arXiv:2008.08294 [cs]}, Aug. 2020.
\newblock arXiv: 2008.08294.

\bibitem{Zhao2021-ca}
He Zhao and Richard~P Wildes.
\newblock Where are you heading? dynamic trajectory prediction with expert goal
  examples.
\newblock In {\em 2021 {IEEE/CVF} International Conference on Computer Vision
  ({ICCV})}, pages 7629--7638. IEEE, Oct. 2021.

\bibitem{zhao2019multi}
Tianyang Zhao, Yifei Xu, Mathew Monfort, Wongun Choi, Chris Baker, Yibiao Zhao,
  Yizhou Wang, and Ying~Nian Wu.
\newblock Multi-agent tensor fusion for contextual trajectory prediction.
\newblock In {\em Proceedings of the IEEE/CVF Conference on Computer Vision and
  Pattern Recognition}, pages 12126--12134, 2019.

\end{thebibliography}
}

\appendix

% \section{Joint Optimization of other SOTA methods}
% We test the applicability of the joint loss term on other SOTA methods: View Vertically~\cite{Wong2022-kx} and MemoNet~\cite{Xu2022-ki}. Our results show that perform equ

% \section{Worst-case Analysis}

% \section{Baseline Comparisons by Interaction Categories}
% Baseline performance across interaction category cross-sections in ETH / UCY and SDD can be seen in Table \ref{table:baseline_int_cats}. We note several analyses: 
% \input{tables/baseline_int_cats}

\section{Additional Trajectory Visualizations across Methods}

\begin{figure*}[t]
\begin{center}
  \includegraphics[width=\textwidth]{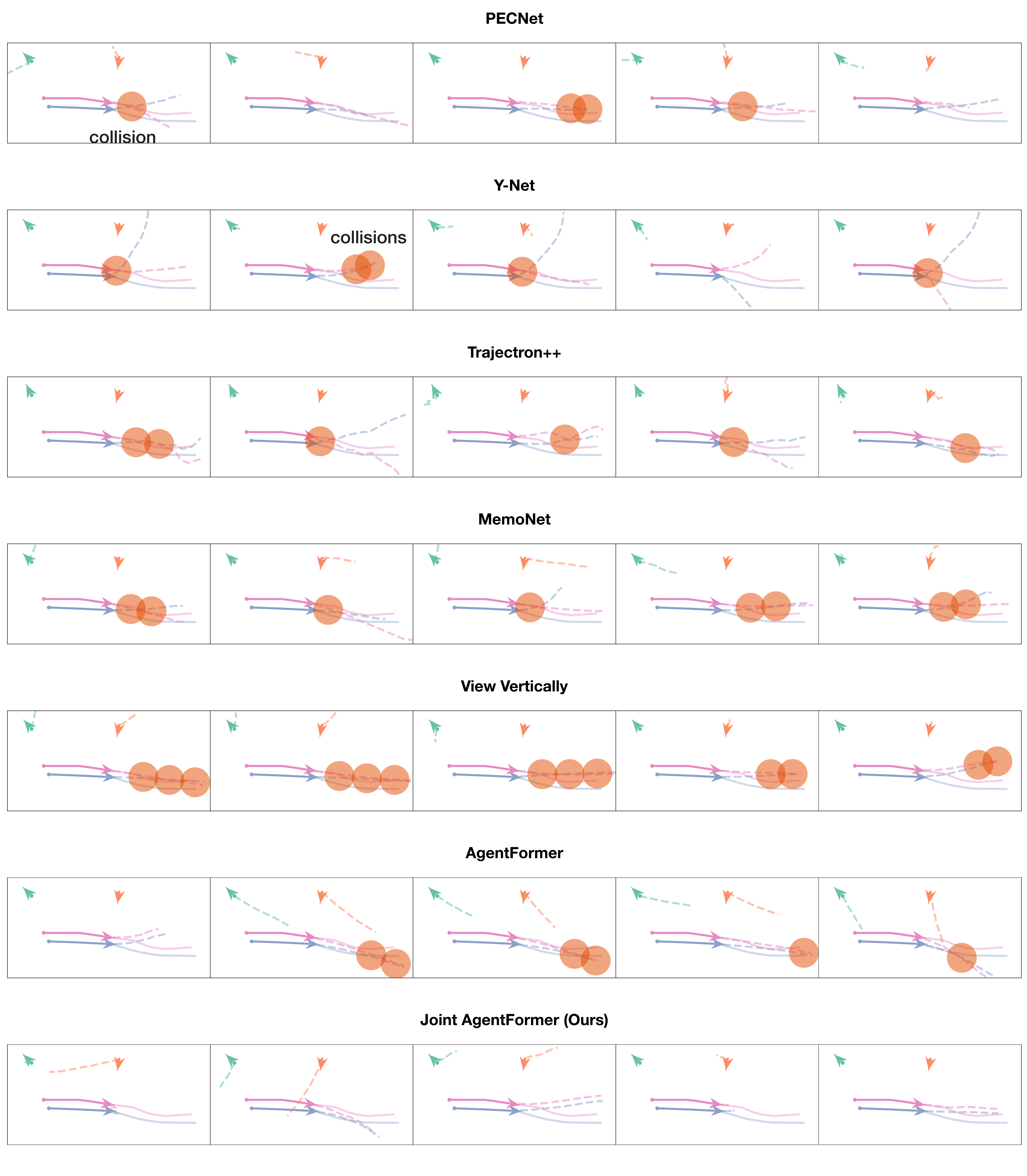}
\end{center}
\caption{Our method exhibits improved collision-avoidance over baselines. While other methods (first 6 rows), including AgentFormer, predicts trajectories with collisions (denoted by orange circles), our method (last row) predicts trajectories with no collisions.}
\label{fig:traj_viz_start}
\end{figure*}

\begin{figure*}[t]
\begin{center}
  \includegraphics[width=\textwidth]{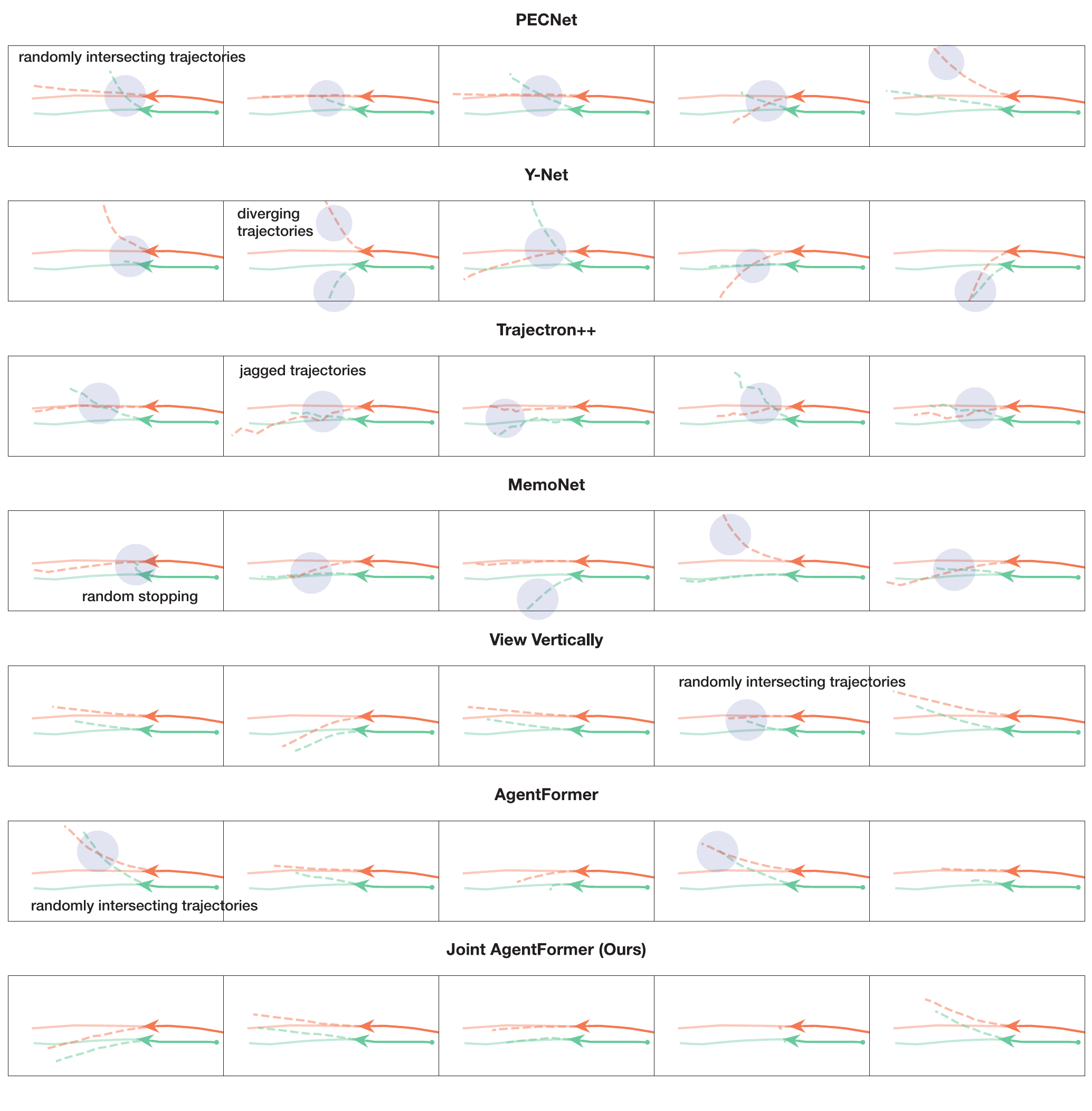}
\end{center}
\caption{Our method exhibits more natural grouping behavior over baselines. While other methods (first rows) predict unnatural trajectories such as diverging or intersecting trajectories (denoted by blue circles) for agents who are obviously grouped, our method (last row), optimized for joint metrics, predicts reasonable grouping behavior while still maintaining diversity of predictions.}
\label{fig:traj_viz_group}
\end{figure*}

\begin{figure}[t]
\begin{center}
  \includegraphics[width=\columnwidth]{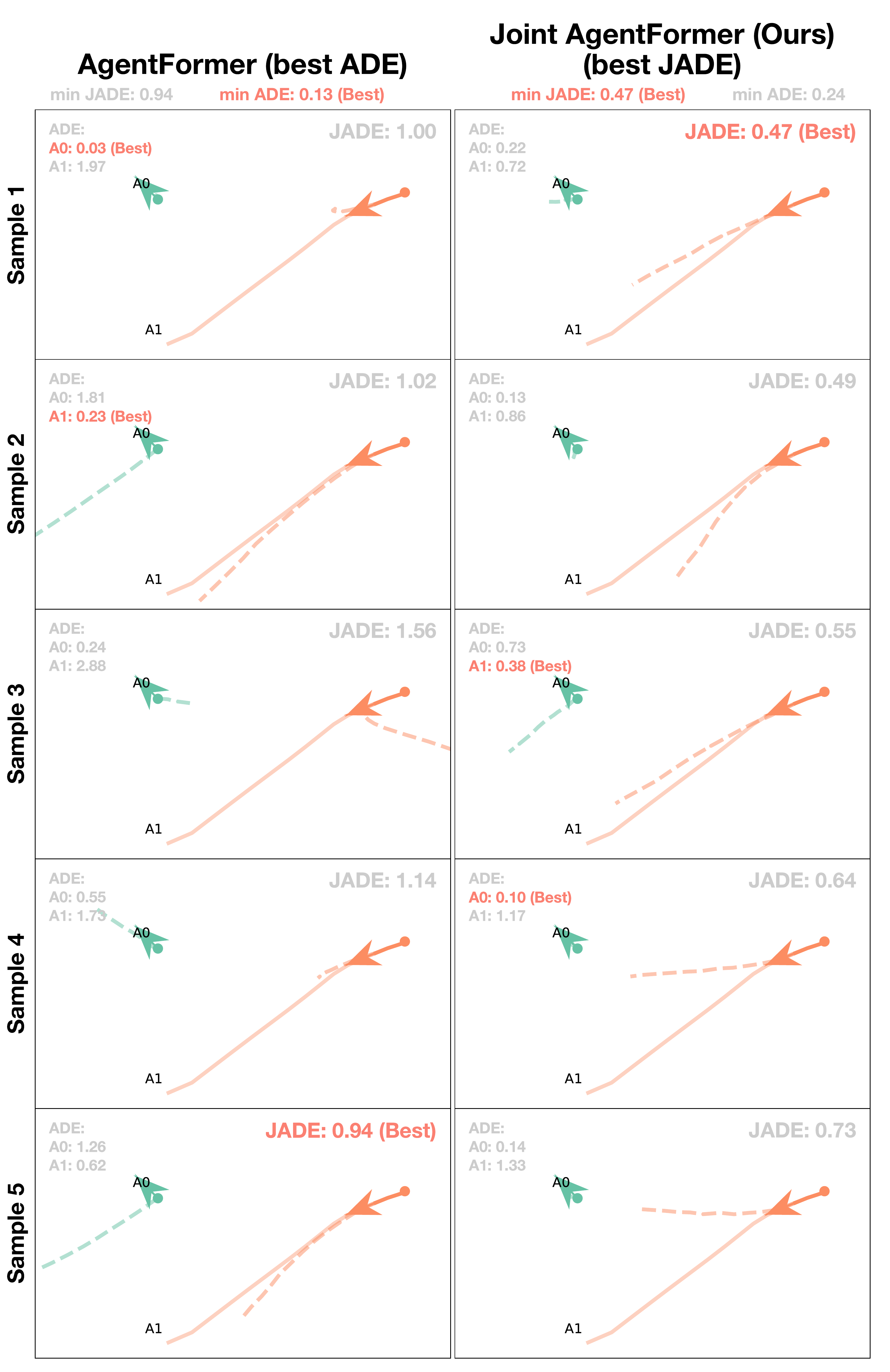}
\end{center}
\caption{AgentFormer vs. Our Method. Of particular interest is qualitative comparison between AgentFormer and our method, as our method was built on AgentFormer. In our qualitative observations, AgentFormer heavily represents the future mode in which a static pedestrian suddenly begins moving at regular speed. This static-to-moving mode is not very likely; static-to-moving pedestrians account for only $1887 / 34161 \approx 5.5\%$ of pedestrians in the ETH / UCY datasets. 
% However, as seen in the Figure, AgentFormer represents this future X times among $20$ prediction samples. 
The reason AgentFormer does this is because under marginal loss, over-representing unlikely modes is not penalized. Marginal optimization favors exploration of unlikely modes over exploitation of the most probable modes, because marginal evaluation assumes that any of the $20^N$ possible ways to mix-and-match $20$ trajectory predictions for $N$ agents are valid. Thus, marginal evaluation favors prediction diversity at the expense of realism. 
On the other hand, under joint loss optimization, our method is allowed not $20^N$ but rather only $20$ ``tries" to generated an accurate joint prediction. Not having the luxury to explore, our method reduces the representation of the unlikely static-to-moving mode. Thus, even though AgentFormer outperforms our method with respect to \textit{ADE}, our method, which outperforms AgentFormer with respect to \textit{JADE}, produces a more realistic set of predictions.}
\label{fig:traj_viz4}
\end{figure}

\begin{figure}[t]
\begin{center}
  \includegraphics[width=\columnwidth]{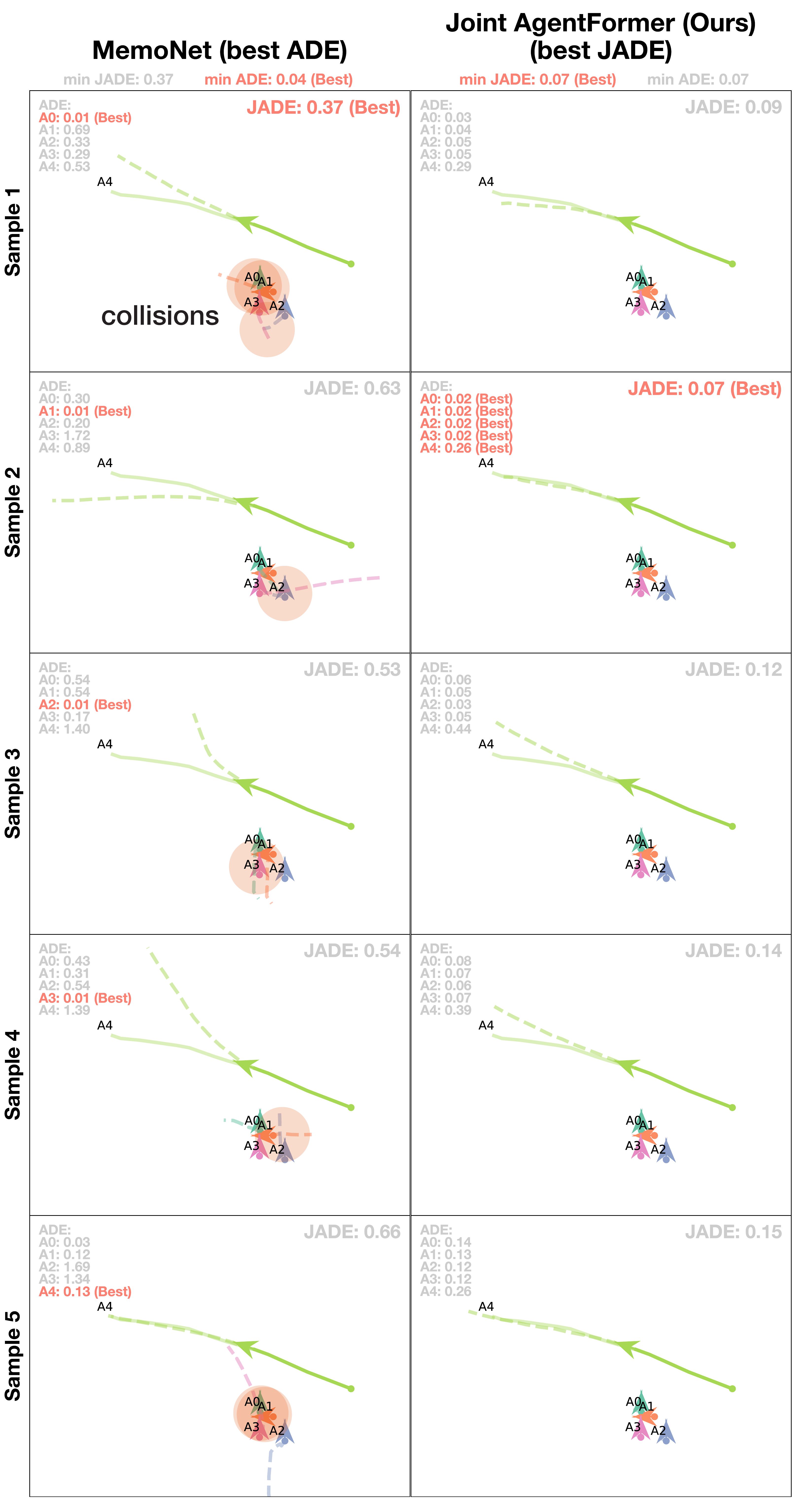}
\end{center}
\caption{Our method outperforms MemoNet, the SOTA with respect to \textit{XDE}, at producing collision-free joint predictions. Although MemoNet \cite{Xu2022-ki} is SOTA with respect to \ADE / \FDE on ETH / UCY Average, it appears to possess minimal social awareness, producing predictions full of collisions (denoted by red circles). 
Furthermore, 
% just as we observe in Figure \ref{fig:traj_viz5}, 
in this Figure we also observe that MemoNet produces unnatural and inconsistent trajectories. For example, the group of static pedestrians in the lower-right corner is expected to maintain its static position, or perhaps all begin moving as a group. However, MemoNet occasionally predicts them moving off in different directions. Thus, even though MemoNet outperforms our method at \textit{ADE}; our method, which outperforms with respet to \textit{JXDE}, generates more natural and consistent predictions.}
\label{fig:traj_viz6}
\end{figure}

\begin{figure}[t]
\begin{center}
  \includegraphics[width=\columnwidth]{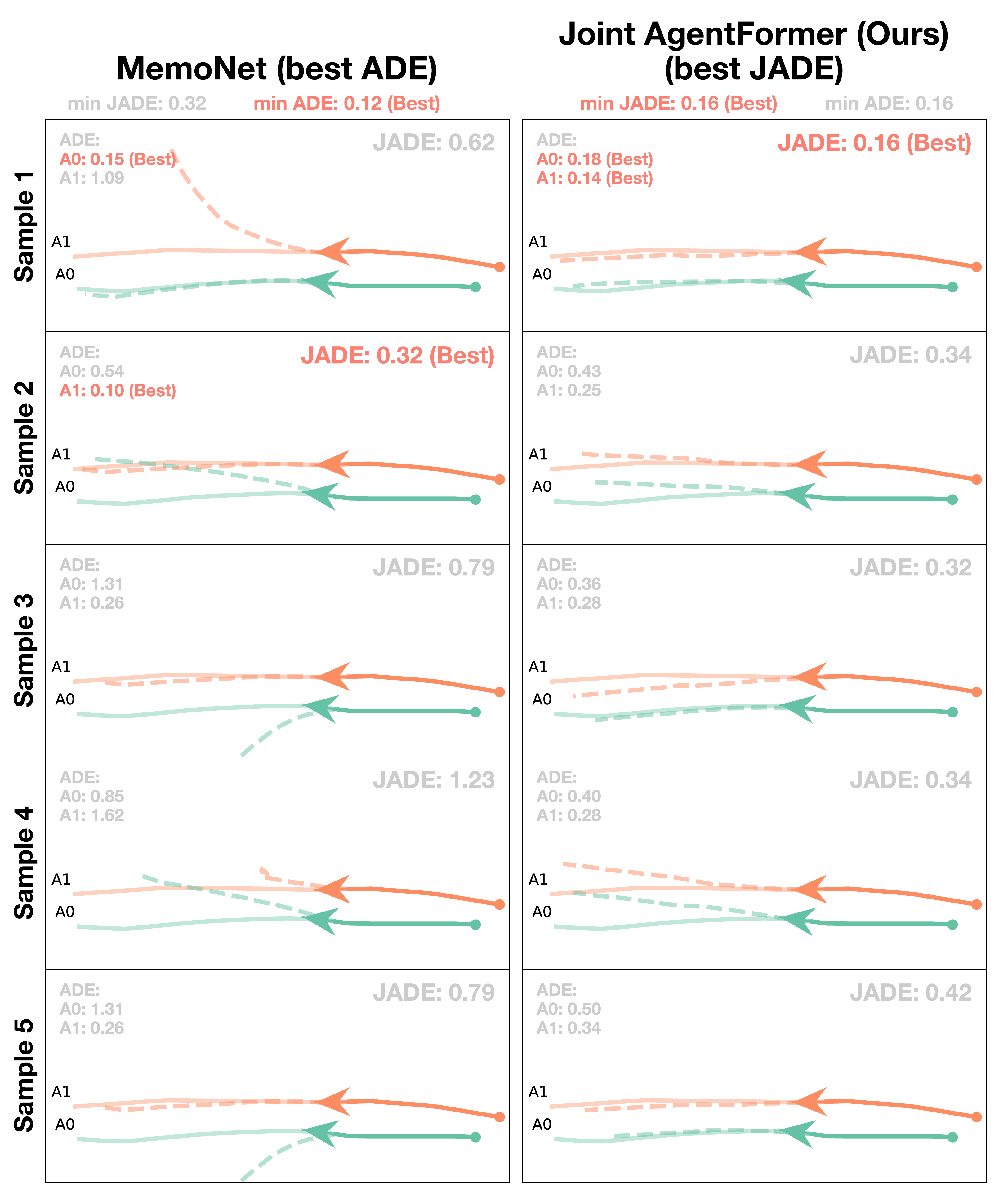}
\end{center}
\caption{Our method outperforms MemoNet, the SOTA with respect to \textit{XDE}, at producing natural and consistent joint predictions. Although MemoNet \cite{Xu2022-ki} is SOTA with respect to \ADE / \FDE on ETH / UCY Average, it appears to possess minimal social awareness, producing predictions with clearly grouped agents going off in different directions. Thus, even though MemoNet outperforms our method at \textit{ADE}; our method, which outperforms with respet to \textit{JXDE}, generates more natural and consistent predictions.}
\label{fig:traj_viz5}
\end{figure}

\begin{figure}[t]
\begin{center}
  \includegraphics[width=.5\columnwidth]{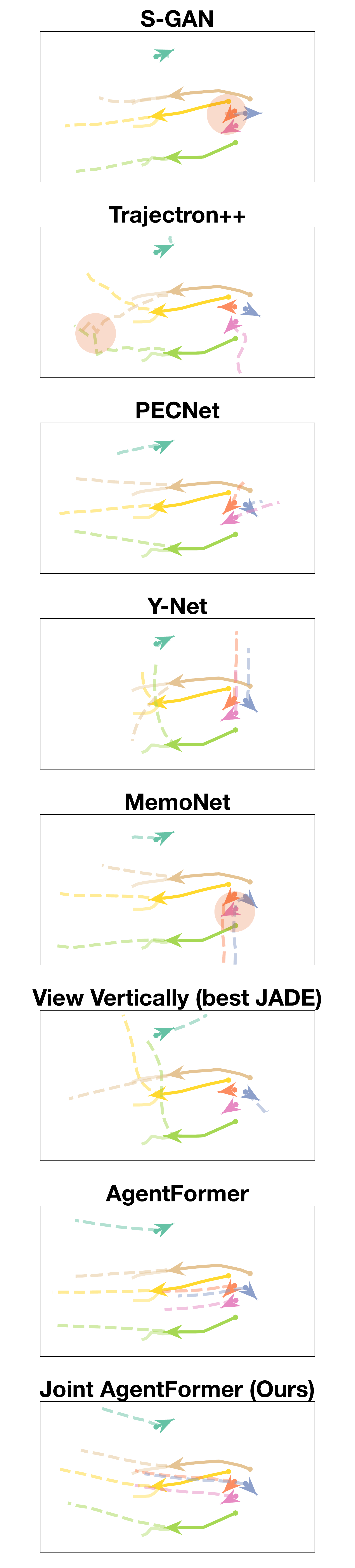}
\end{center}
\caption{Worst-case Analysis. For the sequence shown in this figure, we select the max \JADE example for each method. Even the worst-\JADE prediction of our method produces natural-looking trajectories, while other methods produce collisions (S-GAN, Trajectron++, MemoNet) or inconsistent and random predictions (Y-Net, View Vertically). }
\label{fig:traj_viz_end}
\end{figure}

Additional examples in the style of Figure \ref{fig:traj_viz} of performance comparisons across methods on specific trajectory sequences can be seen in Figures 
\ref{fig:traj_viz_start}-\ref{fig:traj_viz_end}.

\begin{figure*}[t]
\begin{center}
  \includegraphics[width=.9\textwidth]{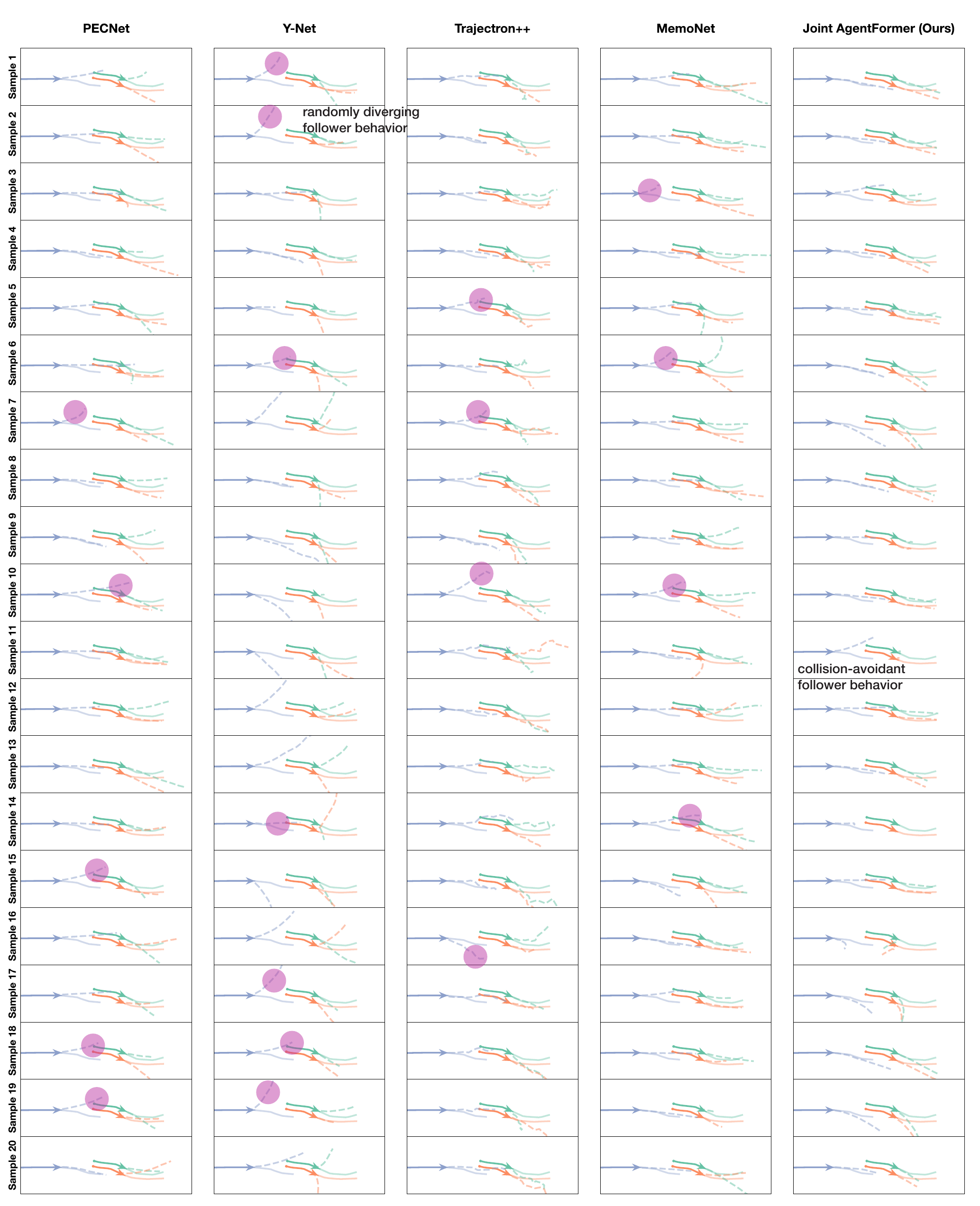}
\end{center}
\caption{Our method exhibits more natural leader-follower behavior over baselines. Other methods (first four columns) predict unnatural trajectories such as sudden direction changes (denoted by purple circles) for the blue agent, who is clearly following along the natural path defined by the orange and green agent. On the other hand, our method (last column) predicts natural leader-follower behavior for the blue agent, while still maintaining visible diversity of predictions. Also of interest is our method's superiority with respect to predicting collision-avoidance as well as natural grouping behavior of the orange and green agents, but we leave it unmarked in this visual to leave the reader's attention upon leader-follower behavior.}
\end{figure*}

\end{document}